\documentclass{article}

\usepackage{arxiv}

\usepackage[utf8]{inputenc} 
\usepackage[T1]{fontenc}    
\usepackage{hyperref}       
\usepackage{url}            
\usepackage{booktabs}       
\usepackage{amsfonts}       
\usepackage{nicefrac}       
\usepackage{microtype}      
\usepackage{lipsum}
\usepackage{amsmath}
\usepackage{amssymb}
\usepackage{todonotes}
\usepackage{graphicx}
\usepackage[numbers]{natbib}
\usepackage{algpseudocode} 
\usepackage[ruled,vlined,linesnumbered]{algorithm2e} 

\title{Natural Language Understanding for Argumentative Dialogue Systems in the Opinion Building Domain}

\author{
  Waheed Ahmed Abro\thanks{Corresponding Author.} \\
  School of Computer Science and Engineering \\
   Southeast University, Nanjing, China  \\
   \& \\
   Institute of Communications Engineering\\
   Ulm University, Germany \\
  \texttt{waheed@seu.edu.cn} \\
  \And
  Annalena Aicher\\
  Institute of Communications Engineering\\
   Ulm University, Germany \\
  \texttt{annalena.aicher@uni-ulm.de} \\
  \And
  Niklas Rach\\
  Institute of Communications Engineering\\
   Ulm University, Germany \\
  \texttt{niklas.rach@uni-ulm.de} \\
  \And
  Stefan Ultes\\
  Mercedes-Benz Research \& Development\\
   Sindelfingen, Germany\\
  \texttt{stefan.ultes@daimler.com} \\
  \And
  Wolfgang Minker\\
  Institute of Communications Engineering\\
   Ulm University, Germany
  \texttt{wolfgang.minker@uni-ulm.de} \\
  \And
  Guilin Qi \\
  School of Computer Science and Engineering\\
   Southeast University, Nanjing, China \\
  \texttt{gqi@seu.edu.cn} \\
}

\begin{document}
\maketitle

\begin{abstract}
This paper introduces a natural language understanding (NLU) framework for argumentative dialogue systems in the information-seeking and opinion building domain. The proposed framework consists of two sub-models, namely intent classifier and argument similarity. Intent classifier model stack BiLSTM with attention mechanism on top of pre-trained BERT model and fine-tune the model for recognizing the user intent, whereas argument similarity model employs BERT+BiLSTM for identifying system arguments the user refers to in his or her natural language utterances. Our model is evaluated in an argumentative dialogue system that engages the user to inform him-/herself about a controversial topic by exploring pro and con arguments and build his/her opinion towards the topic. In order to evaluate the proposed approach, we collect user utterances for the interaction with the respective system labeling intent and referenced argument in an extensive online study. The data collection includes multiple topics and two different user types (native English speakers from the UK and non-native English speakers from China). Additionally, we evaluate the proposed intent classifier and argument similarity models separately on the publicly available Banking77 and STS benchmark datasets. The evaluation indicates a clear advantage of the utilized techniques over baseline approaches on several datasets, as well as the robustness of the proposed approach against new topics and different language proficiency as well as the cultural background of the user. Furthermore, results show that our intent classifier model outperforms DIET, DistillBERT, and BERT fine-tuned models in few-shot setups (i.e., with 10, 20, or 30 labeled examples per intent) and full data setup.

\end{abstract}


\section{Introduction}

The vast amount of often contradicting information in online sources has raised the need for technologies and applications that assist humans in processing and evaluating them. While recent developments in the field of argument mining~\cite{Lawrence2020} provided the tools to automatically retrieve and structure such information from various sources, the resulting data structures are still large and not necessarily intuitive to humans. Argumentative dialogue systems and conversational agents on the other hand can process these structures~\cite{niklas2018EVA, Sakai2020} and provide a natural and intuitive interface to the data. However, the capabilities of such systems are limited by their ability to understand and process user responses to the presented arguments, especially if they are presented by means of natural language. 

Within this work we introduce a natural language understanding (NLU) approach for the information seeking and opinion building scenario discussed above that extracts the required information from the user input: 
\begin{itemize} 
    \item[a)] Direct user commands to the system (for example end discussion, provide more information)
    \item[b)] System arguments referenced in the user utterance,
    \item[c)] User sentiment on the referenced system arguments.
\end{itemize}

In order to do so, we first recognize the general user intent, i.e. the type of utterance or speech act. In a second step, we use semantic similarity measures to identify the argument that the utterance refers to (if this is required). 

Natural language understanding in the argumentation domain in general is quite challenging~\cite{hunter2019} which can be attributed to the complexity of the domain and the comparatively small amount of conversational training data~\cite{shigehalli2020nlu}. Consequently, NLU components in argumentative systems often suffer from small and/or domain-specific training data that hinders the generalization capability. 

One main challenge is to design an NLU component that works in low-data scenarios where only several examples are available per system-specific intent (i.e., so-called few-shot learning setups).
Recently language models, such as Bidirectional Encoder Representations from Transformers (BERT )~\cite{devlin2019bert} trained on large-scale unlabeled corpora have achieved state-of-the-art performance on natural language processing tasks after fine-tuning. These large-scale pre-trained language models generate contextualized word embeddings and also encodes transferable linguistic features such as parts of speech and syntactic chunks~\cite{liu2019linguistic}.

Considering the benefits of the pre-trained language models, we have utilized the BERT model for the herein discussed NLU approach. More precisely, we have fine-tuned BERT for two NLU tasks, namely intent classification and argument similarity. The proposed intent classifier model stack Bidirectional-LSTM (BiLSTM) with an attention mechanism on top of pre-trained BERT model and fine-tune the whole model. However, fine-tuning BERT on a small dataset may result in overfitting which leads to performance degradation. Therefore, the proposed intent classifier combines sentence representation from the argument similarity model with the representation of BERT+ BiLSTM to improve intent classification performance on a small dataset and few-shot setups. To detect the argument that the user refers to his/her utterance from the set of provided arguments, we use the argument similarity model. The argument similarity model fine-tunes the BERT model on a large supervised semantic textual similarity (STS)~\cite{Cer2017STS} benchmark dataset for textual similarity task. Furthermore, the argument similarity model combines contextualized word features from the BERT layer and word features with common sense knowledge obtained from training BiLSTM on top of ConceptNet Numberbatch~\cite{speer2017conceptnet} to produce high-quality sentence representations which can be compared using cosine similarity.
We test the presented approach in the argumentative dialogue system BEA ('building engaging argumentation’)~\cite{bea2019} that assists the user in building an opinion on a specific topic by providing incremental information and tracking the preference of the user towards it. In order to train and evaluate the proposed model, we collect user utterances
labeled with intent and referenced arguments for the interaction with the BEA system for three different topics in an extensive user study. The participants of study were asked to provide user utterances occurring in the interaction with BEA. The data collected from online study form our User Study dataset. Section~\ref{sec:data} discusses data collection for the User Study dataset. Apart from testing our model in the BEA system, we evaluate the proposed intent classifier and argument similarity models on the publicly available Banking77 and STS benchmark datasets. The results are used to evaluate our approach in four different categories:
\begin{itemize}

\item[a)] We compare intent classifier and argument similarity models separately to baseline approaches on the User Study, Banking77, and STS benchmark datasets to test the robustness of the proposed model on different domains. 

\item[b)] We look at a few-shot intent classification scenario where only 10, 20, or 30 training examples for each intent are sampled from full training data to test model performance in absence of sizeable training data.
\item[c)] We train and evaluate the complete pipeline (intent classifier and argument similarity) on separate topics of the User Study dataset to assess the robustness of the model against topic changes.
\item[d)] We collect a separate test data set from non-native speakers with a different cultural background (Chinese students) in order to test the robustness of the proposed model against different levels of language proficiency and cultural diversity.

\end{itemize}

The experimental results show a clear advantage of our proposed approach over the baselines for the intent classification and argument similarity tasks on different datasets. Moreover, the outcomes indicate a high and stable performance of the model for data from topics unseen during training and different language proficiency. 

The remainder of this paper is as follows: Section~\ref{sec:RelWork} introduces related work from the field of natural language processing and argumentative dialogue systems. The proposed NLU framework is discussed in Section~\ref{sec:NLU} together with the utilized dialogue system. In Section~\ref{sec:data}, we report the collection of training and evaluation data. Section~\ref{sec:ExpSetup} discusses the experimental setup and training setting. The results analysis, impact of hyper-parameters, and ablation study are covered in Section~\ref{sec:EvalRes}, followed by a discussion of our findings and a conclusion in Section~\ref{sec:Conc}.

\section{Related Work}~\label{sec:RelWork}
This section covers related work from two separate research fields. We start by giving an overview over natural language processing approaches that deal with intent classification and textual similarity. Subsequently, we provide an overview over existing argumentative dialogue systems and emphasize the respective approach for recognizing user input.

\subsection{Intent Classification and Textual Similarity} 
Previous works on intent classification use deep learning models such as convolutional neural networks~\cite{kim2014cnn, Zhang2015} and recurrent neural networks~\cite{Ravuri2015, Abro2019Intent} to encode a sentence into a fixed-sized vector representation. A classifier is then applied on top of this representation to classify user intent. Joint modeling of intent classification and slot filling improved performance of intent detection~\cite{liu2016attention,ABRO2020KBS}. Furthermore, Goo et al.~\cite{goo2018slot} proposed a slot-gated mechanism that leverages the intent context vector for modeling slot-intent relationships to improve semantic frame results. On the contrary, Bunk et al.~\cite{bunk2020Diet} proposed DIET (Dual Intent and Entity Transformer) model, which obtains dense features from pre-trained word embedding models such as BERT~\cite{devlin2019bert}, ConveRT~\cite{convert-2020}, and combine these with sparse word features. These features are then used by a 2-layer transformer for detecting user intent. In this connection, Casanueva et al.~\cite{dual-encoder-2020} proposed dual sentence encoders and employed pre-trained Universal Sentence Encoder (USE) ~\cite{cer2018universal} and Conversational Representations from Transformers (ConveRT) ~\cite{convert-2020} for predicting user intent. Furthermore, Minaee et al.~\cite{Minaee2021} provides a detailed survey of the pre-trained models for text classification. The survey presents a quantitative analysis of deep learning models performance on text classification benchmarks. Ali et al.~\cite{ali2020graph,ali2020paper} introduced network embedding on weighted probabilistic models for producing personalized paper recommendations. Recently, attention based deep neural models~\cite{Akhtar2020intense,BASIRI2021279,Cambria2020SenticNet} produce state-of-the-art results on sentiment and emotions intensities classification tasks. 


For the textual similarity task, Skip-Thoughts~\cite{Skip-Thought2015} model was proposed which extends the word2vec~\cite{word2vec2013} skip-gram approach from the word-level to sentence-level. The Skip-Thoughts model employs encoder-decoder architecture to learn vector representation of the sentences.
Similarly, FastSent \cite{hill-2016-learning} was proposed which replaced the RNN encoder with word embedding summation of the skip-thoughts model. In contrast, InferSent~\cite{conneau2017InferSent} model utilizes supervised data of the Stanford Natural Language Inference (SNLI) dataset to train a siamese BiLSTM network with max-pooling over the output. The InferSent model outperforms unsupervised methods like Skip-Thoughts. The Universal Sentence Encoder (USE)~\cite{cer2018universal} model extends the InferSent model by training transformer architecture and augmenting unsupervised learning with supervised training objectives. 

In parallel, pre-trained models such as ELMo~\cite{Peters2018}, GPT ~\cite{radford2018improving}, BERT ~\cite{devlin2019bert}, XLNet ~\cite{yang2019xlnet}, ERNIE ~\cite{sun2019ernie}, and MT-DNN ~\cite{liu2019mt-dnn}  trained on large unsupervised corpora, shown significant improvement for intent classification, semantic textual similarity, and various natural language understanding tasks \cite{wang2018glue}. Such pre-trained models allow the downstream task model to be fine-tuned without training from scratch. Furthermore, fine-tuning of pre-trained models produced state-of-the-art results for text classification~\cite{howard2018,Sun2019} and textual similarity~\cite{Reimers2019SBERT,SKBERT}. Inspired by the recent works, we have employed these pre-trained models for the natural language understanding of the arguments.

\subsection{Argumentative Dialogue Systems}
A variety of different argumentative dialogue systems have been introduced in the past years. In~\cite{yuan2008} a dialogue system to enable a computer to engage its users in debate on a controversial issue was introduced. In~\cite{bea2019} the BEA system is proposed which is an argumentative dialogue system that helps a user to form his or her opinion on a certain topic by providing arguments in a spoken dialogue. Rach et al.~\cite{niklas2018EVA} proposed EVA System to discuss controversial topics with the users. Hunter~\cite{Hunter2018APS} discusses formal models of dialogues involving arguments and counterarguments of user models, and strategies for automated persuasion systems (APS). Ma et al.~\cite{MA2020EDS} provides a review of empathetic dialogue systems that respond to users in an empathetic way.


On the other hand, argumentative systems that include natural language input like the IBM Debater\footnote{\url{https://www.research.ibm.com/artificial-intelligence/project-debater/}} are mainly focused on the exchange of arguments in competitive setups like debates, discussions or persuasion~\cite{shigehalli2020nlu, Rakshit2019, nguyen2018dave, Higashinaka2017}. For instance, Rosenfeld and Kraus~\cite{rosenfeld2016persuasion} proposed a methodology for persuading people through argumentative dialogues to invoke an attitude and behavior change. In contrast, we look at a cooperative setup and focus explicitly on the exchange of information and opinions \emph{about} the arguments presented by the system rather than a recognition of user arguments.

\begin{figure*}
    \centering
     \includegraphics{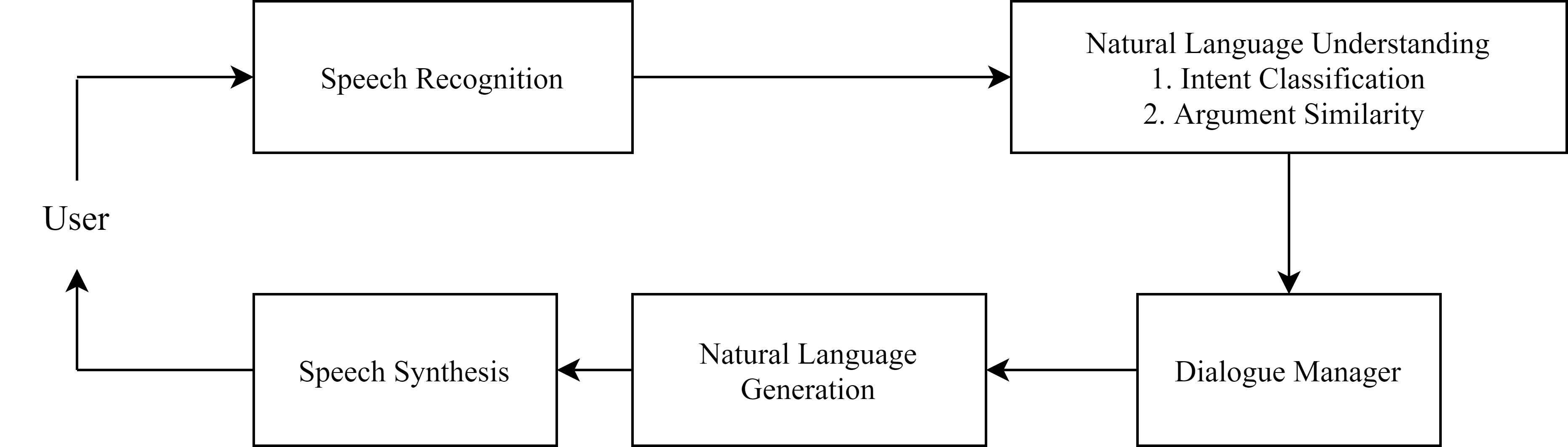}
    \caption{Architecture of a spoken argumentative dialogue system.}
     \label{fig_main}
\end{figure*}

\section{Natural Language Understanding Framework}~\label{sec:NLU}
As shown in Fig.~\ref{fig_main}, the architecture of a spoken human-computer dialogue system is composed of a chain of five modules: \emph{speech recognition, natural language understanding, dialogue manager, natural language generation} and \emph{speech synthesis}. In the following, we will focus on the second module, the Natural Language Understanding. It processes text from the lexical and syntactic levels, converting it into dialogue action at the discourse level. Therefore in general an intent recognition takes place and potential additional information is analyzed. In our setup, the identified user intents are referred to as \emph{speech acts}, which are determined by the used dialogue model. In order to understand and process the user utterance correctly, \emph{what} the user wants (intent classification) and \emph{which argument} the user refers to (argument similarity) have to be identified. Thus, the herein presented NLU framework consists of two components, an intent classifier model and an argument similarity model. The user intent might consist of direct commands to the system (like ``End the conversation") but can also include expressed opinions, sentiments, or preferences towards arguments presented by the system.
Given a successful recognition of the user intent, the referred content of the latter has to be identified. This is accomplished by comparing the user utterance to the known arguments by the means of semantic similarity measures. Further details of both NLU models, as well as the dialogue system utilized throughout this work are given in the following sections.

\subsection{BEA: An Application Scenario}
To test the herein described NLU framework we make use of the argumentative dialogue system BEA~\cite{bea2019} as an application scenario. BEA incrementally presents arguments on a controversial topic, allows users to express preferences towards and between these arguments and utilizes the responses to model and monitor the user opinion in view of the discussed topic throughout the interaction. The goal of BEA is to engage the users into an intuitive and natural dialogue allowing them to explore different arguments with diverging stances and various subtopics. In contrast to competitive systems BEA does not pursue a persuasive approach but tries to provide pro and con aspects on a controversial topic to help the user to build a balanced opinion.\\
In order to navigate through a large amount of arguments and divide the discussed information in reasonable and logically consistent parts, the system utilizes an argument structure based on the argument annotation scheme introduced by Stab et al.~\cite{stab2014annotating}. This scheme was originally introduced for annotating argumentative discourse structures and relations in persuasive essays and meets our purpose to offer the user a fair chance to decide unprejudiced which side (pro/con) to prefer or reject. According to Stab et al. an argument consists of several argument components (\textit{major claim}, \textit{claim} and \textit{premise}) and two relations (\emph{support} and \emph{attack}) between them. Usually a single \textit{major claim} formulates the overall topic of the debate, representing the root node in the tree graph structure.
 Likewise to Aicher et al.~\cite{bea2019} in the following we use sample debate from the \textit{Debatabase} of the idebate.org\footnote{https://idebate.org/debatabase (last accessed 16 September 2021).\\Material reproduced from www.iedebate.org with the permission of the International Debating Education Association. Copyright \copyright~2005 International Debate Education Association. All Rights
Reserved.} website with the \textit{major claim} \textit{"Marriage is an outdated institution''}.\\ \textit{Claims} are allegations which formulate a certain opinion targeting the \textit{major claim} but still need to be justified by further arguments, \textit{premises} respectively. Hence, a \textit{claim} (parent node) is either supported or attacked by at least one other premise (child node). For the remainder of this work, we refer to a single node, i.e., an argument component in the structure as \emph{argument}.\\
We only focus on non-cyclic graphs, meaning that each premise only targets one other component, leading to a strictly hierarchical structure. Furthermore, the annotation scheme distinguishes two directed relations a premise can have towards a claim (\textit{support} and \textit{attack}). Between sibling nodes there exists no explicit relation.\\
Due to the generality of the annotation scheme, the system is not restricted to certain data and generally every argument structure that can be mapped into the applied scheme can be processed by the system.\\
In the interaction, BEA introduces sibling arguments related to the same parent argument in the tree simultaneously. In particular all available argument components attacking or supporting the parent node are introduced. Thus, the user is able to express preferences\footnote{For ease of reading, 'preferences' here denotes both a negative (rejecting) and positive (approving/preferring) attitude towards an argument.} between the siblings or navigate to another sub-structure depending on his/her interest. If a user expresses a preference, it is crucial the system can identify the user intent and the sibling which is preferred. This expressed preference is incorporated into a calculation that determines the user's overall opinion on the topic of the discussion and is updated in real-time during the interaction.

\begin{table*}
\centering
\caption{Available speech acts and their corresponding user and system action.}\label{tab:avail_moves}

\begin{tabular}{p{2.4cm}|p{6.2cm}|p{6.2cm}}
\hline
\textbf{Speech Act} &\textbf{User Action} &\textbf{System Action} \\ 
\hline
stance & Request for the current overall stance of the user. & WBAGs are used to calculate and returns current user stance on the overall topic. \\
\hline
exit & Request to end the interaction with the system. & System terminates interaction with complimentary closing. \\
\hline
level-up & Request to return to the previous argument (parent). & Changes the current state and switches to the corresponding parent node (one level up). \\
\hline
why(argument) & Request for further information on an argument. & Provides information about the current argument by introducing all of its child nodes. \\
\hline
prefer(argument) & States preference towards the referenced argument over all its siblings. & Calculates new stance according to the preference model and updates tree.\\
\hline
reject(argument) & Rejects an argument. & The argument and all its corresponding child nodes are rejected, thus, a new stance is calculated according to the preference model and the tree is updated.\\
\hline
\end{tabular}
\end{table*}

The interaction is divided in turns such that each user action ('move') is followed by a system response and vice versa. The six different user moves the user is able to choose from and the corresponding system actions are shown in Table~\ref{tab:avail_moves}. These moves are equivalent to the intents that have to be recognized by the NLU. Before starting the interaction with BEA, these moves are explained in an instruction for the users on how to use BEA.\\ 
Three moves (\textit{prefer}, \textit{reject}, and \textit{why}) refer to a specific argument and require the NLU to identify this argument. Whereas \textit{prefer}, \textit{reject} allow the user to express his or her opinion towards the argument, the \textit{why} move can be used to ask the system for further information on the argument. Thus, the selected argument the \textit{why} move refers to, becomes the parent node and its attacking and supporting children are displayed. Since BEA introduces only siblings related to one parent node at the same time, the list of siblings serves as the list of possible reference arguments for the NLU.

In addition to that, the user is able to request the calculated opinion (weight) on an argument (\textit{stance})\footnote{In case a user has no opinion on the major claim but on certain subtopics, he or she can determine her stance upon the major claim.}. To calculate the user stance the system uses the preference statements and determines the respective stance by utilizing weighted bipolar argumentation graphs (wBAGs)~\cite{WBAG2018}. In order not to go beyond the scope of this work, for the detailed description of the calculation of preference weights and strengths we refer to Aicher et al.~\cite{bea2019}. They describe the iterative propagating through the tree structure and update after every preference statement.
The user can navigate through the tree, by using the \textit{why} move to descend and the \textit{level up} move to ascend. The latter takes him/her back to the previous argument. The conversation with BEA can be terminated by the user at any time by performing the \textit{exit} move.\\
The natural language generation of the system is based on the original textual representation of the argument components. As discussed in ~\cite{bea2019} the annotated sentences were slightly modified to form a stand-alone utterance which serves as a template for the respective system response. Additionally, a list of natural language representations for each type of move was defined. During the generation of the utterances, the explicit formulation and introductory phrase is chosen from this list randomly. E.g. regarding the \textit{why} move the relation (\textit{support} or \textit{attack}) of the presented children is clearly stressed by specific formulations, such as \textit{This claim is supported by the argument that...} and \textit{A contrary indication is the fact that...}.\\
In the following, an exemplary dialogue is shown to demonstrate the targeted user system interaction. In  claim\footnote{For the sake of simplicity we define this argument as the \textit{major claim} of this subdialogue.} \textit{"Marriage undermines same-sex couples and single parent families as legitimate ways of raising children."}, which is addressed by the user with a \textit{why} move.

\begin{itemize}
\item[System:]\textit{"This claim is supported by the argument that marriage is seen as the best way to raise children (C1). Another argument in favor is the existence of marriage is essentially saying that same-sex couples and single parents are less able of raising children than heterosexual couples (C2). A contrary indication is the fact that the idea that the existence of marriage undermines other methods of raising children is ridiculous (C3)."}
\item[User:]\textit{"I do not believe that this is the best way to raise children."}
\item[System:]\textit{"It is interesting that you rejected this claim (C1)."}
\end{itemize}

In order to realize this dialogue the system has to fulfill two tasks. First, the intent of the user has to be identified (rejection) and second, it has to determine which of the three claims the user wants to reject ("marriage is seen as the best way to raise children"). Therefore, the intent classification model addresses the first problem, introduced in Subsection~\ref{sec:intent}. The second part represents the argument similarity model described in Subsection~\ref{sec:sim}. Both taken together form the complete system pipeline.

\begin{figure*}
    \includegraphics[height=.80\linewidth,width=.99\linewidth] {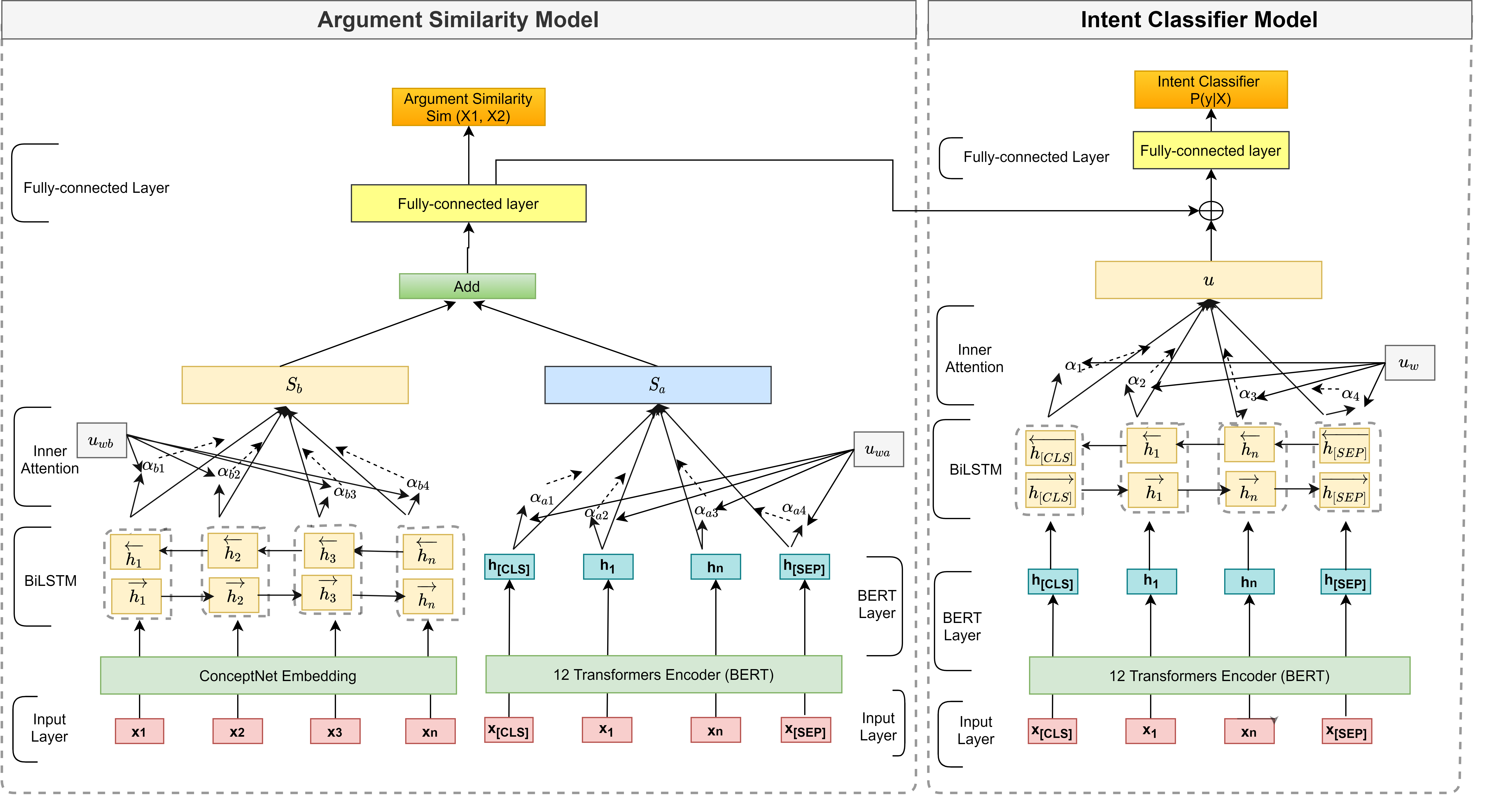}
    \caption{An illustration of argument similarity and intent classifier model. Argument similarity model generates sentence representations $S_a$, $S_b$ by applying inner attention on BERT encoder and BiLSTM encoder, respectively. These representations are added together and pass through a fully-connected layer. Intent classifier model obtains sentence representation $u$ by applying inner attention on BERT+BiLSTM encoder. The final representation is produced by concatenating sentence representation $u$ and $s$ and passing through a fully-connected layer.}
     \label{fig_intent}
\end{figure*}

\subsection{Intent Classifier Model}\label{sec:intent}
The architecture of the proposed intent classifier model is depicted in Fig.~\ref{fig_intent}. The model consists of two main components namely: BERT Transformer Encoder and BiLSTM classifier. 
Further details of these components are given in the following subsections.

\subsubsection{BERT Transformer Encoder}
The BERT consists of several Transformer~\cite{Vaswani2017trans} encoder layers. Each transformer encoder has a multi-head self-attention mechanism and a fully-connected neural network with residual connections. For every word in the user utterance, each head of the self-attention layer computes query, key, and value vectors. 
The outputs of these heads are concatenated and passed through a linear layer to create a weighted representation of each word. 

The BERT model computes user utterance representation as follows: The user utterance is given as input to WordPiece~\cite{wu2016google} tokenizer, which splits the utterance into a list of tokens and then combines token, position, and segment embeddings for producing a fixed-length vector. Moreover,  the special classification \verb|[CLS]| token is added at the start of each utterance. Similarly, special \verb|[SEP]| token is inserted at the end of each sentence as a final token. The BERT encoder then computes the user utterance representations i.e. hidden states for each token $x_{t}$ as shown in equation~\ref{eq_sent_rep}.
\begin{equation}
H_t^n = BERT(x_{[CLS]},x_{1},..,x_{t},x_{[SEP]})
\label{eq_sent_rep}
\end{equation}
where $H_t^n = (h_{[CLS]}^n,h_{1}^n,..,h_{t}^n,h_{[SEP]}^n)$, n denotes number of BERT encoder layer, and $h_t$ is the contextual representation of token $t$. The final hidden state, $H_t^{N} (N=12) $ of each token is passed to a task-specific LSTM layer.

\subsubsection{Bidirectional-LSTM on BERT}
The Long short-term memory (LSTM)~\cite{hochreiter1997long} is a powerful architecture capable of capturing long-range dependencies via time-connection feedback. Our proposed model stacks a Bidirectional-LSTM (BiLSTM) on top of the final BERT encoder layer. The forward LSTM and a backward LSTM of BiLSTM read in the final hidden states of all the words $H_t^{N} (N=12)$, produced by BERT in two opposite directions and generates output sequences $\overrightarrow{h_t}$ and $\overleftarrow{h_t}$. The two outputs are then concatenated to access both past and future context for a given word as given in equation~\ref{eq_lstm}.

\begin{equation}
\begin{gathered}
\overrightarrow{h_t} = \overrightarrow{LSTM}(H_1^{N},...,H_t^{N})
\\
\overleftarrow{h_t} = \overleftarrow{LSTM}(H_1^{N},...,H_t^{N}) 
\\
h_t = [\overrightarrow{h_t},\overleftarrow{h_t}]
\label{eq_lstm}
\end{gathered}
\end{equation} 

where $h_t$ is the representation of the given word obtained by concatenating the forward hidden state $\overrightarrow{h_t}$ and backward hidden state $\overleftarrow{h_t}$.

\subsubsection{Inner-Attention}

To encode the variable-length sentence into a fixed-sized vector representation, we employ an attention mechanism.  The attention mechanism is used to give more focus on the important information of the sentence. The attention mechanism is applied to the whole hidden states $H = (h_1, h_2, ..., h_t)$ of BiLSTM to generate vector representation of the sentence. As suggested in~\cite{lin2017structured}, we use multiple attention views to focus on a different part of the sentence. The attention mechanism is defined as follows:

\begin{equation}
\begin{gathered}
u_w = W_2 (tanh(W_1 H^{\intercal} + b_1) + b_2  \\
\alpha  = softmax (u_w)
\\
u = H \alpha^{\intercal}
\label{eq_attention}
\end{gathered}
\end{equation}

where $W_1 \in \mathbb{R}^{d_{w} \times 2d}$ and $ W_2 \in \mathbb{R}^{r \times d_w}$ are the trainable weight metrics; $b_1 \in \mathbb{R}^{d_w}$ and $b_2 \in \mathbb{R}^{r}$ are the trainable bias, here $r$ is the number of attention heads, $d$ represents the number of hidden units of LSTM, $d_{w}$ is the size of vector parameters we can set arbitrarily. The size of $H$ is  $\mathbb{R}^{T \times 2d}$ and the attention matrix $\alpha$ size is $\mathbb{R}^{T \times 2d}$.

The softmax function is applied along the second dimension of its input, which ensures computed weights sum up to 1. We then compute the weighted average of context vectors $r$ by multiplying the attention matrix $\alpha$ and LSTM hidden states $H$ to generate the sentence representation $u$. The final representation $u$ along with sentence representation $s$ obtained from the argument similarity model is given as input to a fully connected layer for predicting the corresponding user intent (speech act).
       

\begin{equation}
\begin{gathered}
\hat{y} = \text{softmax}(W [u, s] + b)
\label{eq_inten_final}
\end{gathered}
\end{equation}
where $\text{softmax}(z_i) =\frac{e^{z_{i}}}{\sum_{k}e^{z_{k}}}$, $W \in \mathbb{R}^{d\times |I|}$ and $b \in \mathbb{R}^{|I|}$ are the weights and bias of the fully connected layer respectively, s is the sentence representation as defined in Eq~\ref{eq_sim}, and $I$ denotes the user intent vocabulary. Furthermore, the model minimizes the cross-entropy loss between true user intent and predicted user intent $\hat{y}$.
\begin{equation}
    \mathcal{L} = - \sum_{i=1}^l {y}^{i} log(\hat{y}^{i})
\end{equation}
where $i$ is the number of intents while ${y}^{i}$ is actual user intent and $\hat{y}^{i}$ is predicted user intent.

\subsection{Argument Similarity Model}\label{sec:sim}
The argument similarity model operates on the contextual word features obtained from the BERT model and word features with common sense knowledge obtained from training BiLSTM on top of ConceptNet Numberbatch~\cite{speer2017conceptnet}. At a high level, our model consists of two parts: BERT Encoder and BiLSTM.

In the first part, each word of the sentence is passed through BERT encoder layers and output vectors are given as input to the inner-attention layer. Attention mechanism provides summation vectors which are dotted with BERT output vectors to generate a sentence vector. This yields

\begin{equation}
\begin{gathered}
H_a = BERT(x_{[CLS]},x_{1},..,x_{t},x_{[SEP]}) \\
u_{wa} = W_{a2} tanh(W_{a1} H_a + b_{a1}) + b_{a2} \\
\alpha_a  = softmax(u_{wa})
\\
S_a = H_a \alpha_a^{\intercal}
\end{gathered}
\label{eq_bert_enc}
\end{equation} 

where $W_{a1} \in \mathbb{R}^{d_{w} \times d}$ and $ W_{a2} \in \mathbb{R}^{r \times d_w}$ are the weight metrics; $b_{a1} \in \mathbb{R}^{d_w}$ and $b_{a2} \in \mathbb{R}^{r}$ are the bias, where $r$ is the number of attention heads, $d$ represents the size of BERT output vectors. 
We compute weighted average of context vectors $r$ by multiplying the attention matrix $\alpha_a$ and BERT output vectors $H_a$ to generate the sentence representation$ s_{1a}$.

In the second part, each sentence is passed through the ConceptNet Numberbatch embedding layer to obtain a semantic vector for each word. ConceptNet Numberbatch embedding combines embeddings from word2vec~\cite{mikolov2013efficient}, GloVe \cite{pennington2014glove}, and structured knowledge from ConceptNet~\cite{speer2012representing}, which provide common sense knowledge along with surrounding word context.These word embeddings are given as an input to the BiLSTM layer for modeling the temporal relationship between word embeddings. Then, the model utilizes an attention mechanism on the hidden states of BiLSTM to generate a vector representation of the sentence.

\begin{equation}
\begin{gathered}
\overrightarrow{H_b} = \overrightarrow{LSTM}(x_{1},..,x_{t})\\
\overleftarrow{H_b} = \overleftarrow{LSTM}(x_{1},..,x_{t}) 
\\
H_b = [\overrightarrow{H_b},\overleftarrow{H_b}] 
\\
u_{wb} = W_{b2} tanh(W_{b1} H_b + b_{b1}) + b_{b2} \\
\alpha_b  = softmax(u_{wb})
\\
S_b = H_b \alpha_b^{\intercal}
\end{gathered}
\label{eq_blstm_enc}
\end{equation}

where $H_b$ represents all hidden states of BiLSTM; $W_{b1} \in \mathbb{R}^{d_{w} \times 2d}$ and $ W_{b2} \in \mathbb{R}^{r \times d_w}$ are the weight metrics; $b_{b1} \in \mathbb{R}^{d_w}$ and $b_{b2} \in \mathbb{R}^{r}$ are the bias, $d$ represents the number of hidden units of LSTM, $r$ is the number of attention heads. The weighted sum of hidden states based on attention score weights $\alpha_b$ are used to generate final representation $s_b$.
The outputs of two components are added together and passed through the fully-connected layer to generate final sentence embedding as shown in Fig.~\ref{fig_intent} and given in equation~\ref{eq_sim}.

\begin{equation}
s = W(s_a + s_b) + b
\label{eq_sim}
\end{equation} 

where $W$ and $b$ are the weight matrix and bias for a fully-connected layer.

In order to produce semantically meaningful sentence embeddings, we train the argument similarity model on the STS benchmark dataset. The model takes sentence-pair $(s_1, s_2)$ as input. For sentence $s_1$, the model computes two vector representations. The first vector representation $s_a$ is obtained from the BERT encoder part as given in equation~\ref{eq_bert_enc}. The second vector representation $s_b$ is generated by a BiLSTM encoder as shown in equation~\ref{eq_blstm_enc}. The two vector representations are added together and pass through a fully connected layer to produce final representation of the sentence. Similarly for sentence $s_2$, the model calculates sentence embedding with the BERT encoder part and BiLSTM part. The sentence embeddings generated by each part are added together and passed through a fully connected layer. Finally, the embeddings of $s_1, s_2$ are compared using cosine similarity. The model minimizes the mean squared error between the predicted cosine similarity score and the labeled similarity score. We choose the STS dataset for training the model as it fits best for the argument similarity task of determining the similarity between two sentences. At inference time, the model generates vector representations for user utterance and possible arguments. Cosine distance between these vectors is then calculated and the closest argument vector to a user utterance vector is identified as a reference argument.

\section{Data Collection}~\label{sec:data}
\noindent In order to evaluate our NLU framework, we collected natural language utterances labeled with intent and (if needed) reference argument in an online survey. To this end, participants were asked to paraphrase possible (pre-defined) user utterances occurring in the interaction with BEA.  
After emphasizing to formulate all answers in their own words and showing four examples, the survey was conducted by showing three random arguments to the participants and asking them to reformulate different requests for each speech act. In the case of \textit{prefer}, \textit{reject}, and \textit{why} it was clearly specified which argument the subjects should refer to. To reduce a potential bias the formulation of the instruction was altered for each speech act and participant. For instance, an instruction was: ``Please formulate that you agree with the argument, that 'nuclear weapons had become a source of extreme risk'\footnote{Copyright IBM 2014. Released under CC-BY-SA.}".
To ensure the quality and validate the answers, copy-pasting and skipping answers were not allowed. Furthermore, we required at least five words in each response, which referred to an argument. To make sure that all participants were paying attention to the instructions, a control question was added which asked the user to precisely repeat a sentence.
The data collection was divided into two parts with separate user groups: In the first part, we conducted an anonymous survey via clickworker\footnote{https://marketplace.clickworker.com (last accessed 06 May 2020)} with 200 native English speakers from the UK to collect data for training and testing. In the second part, the group of participants consisted of 15 Chinese Master and Ph.D. students, to test whether there is a measurable effect if the interaction is conducted with non-native speakers. 
As we aim to evaluate our framework especially with regard to cross-domain applicability, we generated samples on three different topics. Two out of three arguments were taken from an annotated debate on the topic \textit{"Marriage is an outdated institution''} from the \textit{Debatabase} of the idebate.org\footnote{https://idebate.org/debatabase (last accessed 09 January 2018)} website~\cite{niklas2018arg}. The other remaining argument was sampled from the IBM corpus on claim and evidence detection~\cite{aharoni2014claim} for one of the two topics \textit{All nations have a right to nuclear weapons} and \textit{The sale of violent video games to minors}. The data collection resulted in 1616 valid user responses for the first group and 143 responses from the second group. 

\section{Experimental Setup }~\label{sec:ExpSetup}

In this section, we define the experimental setup for evaluating the proposed NLU approach. We evaluate the proposed model with respect to four different categories: The first one evaluates intent classification and argument similarity models separately against suitable baselines on different dataset described below: 

\textbf {User-Study:} In Section~\ref{sec:data}, we have discussed the data collection of the User Study dataset. For the experiments, we divided the dataset into train and test sets. We train the intent classifier model on the two topics i.e., \emph{All nations have a right to nuclear weapons} and \emph{The sale of violent video games to minors}. The model is then evaluated on a large test set concerned with \emph{marriage is an outdated institution}.  We adopt this train and test data split to get an estimate of the model's robustness against topic changes. The statistics of train and test samples for each speech act are given in table~\ref{tab:class_dist} and table~\ref{tab_intent_dataset} provides a sample example of each speech act.

\textbf {BANKING77:} The dataset of Coope et al. ~\cite{CoopeFarghly2020}, dubbed BANKING77, is composed of 13,083 customer service queries annotated with 77 intents. The dataset is divided into train and test sets. The test set contains 3080 examples and the full training set contains 10003 examples. 

\textbf {STS benchmark (STSb):} The STS benchmark~\cite{Cer2017STS} dataset is a popular dataset for training and evaluating textual similarity task. It comprises 8628 sentence pairs from three categories: captions, news, and forums. The dataset is divided into train-set (5749), valid-set (1500), and test-set(1379). We train the argument similarity model on the STSb training set and evaluate its performance on the user study dataset and STSb test dataset by computing cosine-similarity between the sentence embeddings. Furthermore, the user study test dataset comprises 2028 sentence pairs for 501 user utterances. For the user study dataset at prediction time, the model generates vector representations for user utterance and possible arguments. Cosine distance between these vectors is then calculated and the closest argument vector to a user utterance vector is identified as a reference argument.  

In the second evaluation category, we look at a few-shot intent classification scenario where only 10, 20, or 30 training examples are sampled for each intent from full training data to get an estimate of the model's performance when small training data is available. It is noteworthy that our argument similarity model is trained on STSb and doesn’t require task-specific training data whereas the intent classifier model needs task-specific training data to learn the required system-specific intents. Due to this reason, we check few-shot setups for the intent classification task where few task-specific training data is available.

In the third evaluation category, we train and evaluate the complete pipeline (intent classifier and argument similarity) on separate topics of the user study dataset to assess the robustness of the model against topic changes. 

The fourth evaluation category compares the results achieved with utterances from native speakers against results achieved with utterances from non-native speakers with a different cultural background to get an estimate of the model's sensitivity towards language proficiency.

\begin{table}
\begin{center}
\caption{Train and test examples statistics for each speech act of user study dataset.} \label{tab:class_dist}
\begin{tabular}{p{1.8cm}|p{1.2cm}|p{1.2cm}|p{1.2cm}}
\hline
Speech Act & Train & Test (UK) & Test (China) \\
\hline

Exit & 72  & 73 & 16 \\
Level-up & 72  & 73 & 16 \\
Stance & 71 & 74 & 15 \\
Why & 189 & 203 & 32 \\
Prefer & 79  & 305 & 33 \\
Reject & 110 & 304 & 31 \\
\hline
\end{tabular}
\end{center}
\end{table}

\begin{table*}
\centering
\caption{Example utterances with annotated labels from user study dataset.} \label{tab_intent_dataset}
\begin{tabular}{l|l}
\hline
\textbf{Utterance} &\textbf{Label} \\ 

\hline
What is my stance right now?                 & stance move \\
\hline
I would like to finish.				& exit move \\
\hline
Please return to the previous argument. 					& level up move \\
\hline
Please tell me more why marriage promotes better way to raise child. 					& why move \\
\hline
I think marriage is good way to raise children 					& prefer move \\
\hline
I reject argument about marriage is an unreasonable expectation		& reject move \\
\hline
\end{tabular}
\end{table*}

\subsection{Training Setup}
For the intent classification, we employ the Bert-Base model\footnote{\url{https://storage.googleapis.com/bert\_models/2020\_02\_20/uncased\_L-12\_H-768\_A-12.zip}} with 12 Transformer layers, 768 hidden states, and 12 self-attention heads. The size of the hidden units in uni-direction LSTM is 512, inner-attention hidden layer $d_w$ is set to 600, and the number of attention head $r$ is 5. Furthermore, we use Adam optimizer with default values of $\beta_1 = 0.9$ and $\beta_2 = 0.99 $, and a learning rate of $1e-4$ and $2e-5$ for training the BiLSTM and fine-tuning the whole model respectively.
Each update is computed through a batch size of 8 or 16 training examples and the number of epochs per batch are 32, 25, 16, and 8 epochs for 10-shot, 20-shot, 30-shot, and full-data settings, respectively. We apply the dropout as a regularization technique for our model to avoid over-fitting. We apply dropout after output of each BiLSTM layer and output of each sub-layer BERT encoder layers. We set the dropout rate as $0.1$ for all dropout layers. 

We employ the transformers~\cite{Wolf2019HuggingFacesTS} library to train our intent model.

For training argument similarity, we used Adam optimizer with a learning rate of $2e-5$ and a batch size of 16 training samples. Furthermore, the model uses the pre-computed 300-dimensional word embeddings ConceptNet Numberbatch. The number of hidden units in uni-direction LSTM is 512 and the number of attention heads is 5. The model is trained for 8 epochs. 

The evaluation metric used for intent classification is the accuracy metric. For the argument similarity task on the user study dataset, the model performance is measured by the accuracy of identifying user reference arguments.  As suggested in~\cite{reimers-2016-task}, we use Spearman correlation for the semantic textual similarity (STS) task. The Spearman's rank correlation is computed between the cosine-similarity of sentence embeddings and gold labels for the STS dataset.

\subsection{Sequential Training}

The proposed approach consists of two sub-models namely: intent classification and argument similarity. We trained these models in a sequential manner. We first train the argument similarity model on the STS benchmark dataset. The detailed training steps are presented in algorithm~\ref{algo1}. After training the argument similarity model, its weights are fixed. The sentence representation obtained from the argument similarity model is then given as an input to a fully connected layer of intent classifier model. 
The proposed approach then trains the intent classifier model; its training is divided into two stages. In the first stage, we freeze the BERT encoder parameters and only train the task-specific BiLSTM and fully connected layer for four epochs. In the second stage, we unfreeze all BERT encoder parameters and fine-tune all parameters of BERT encoder as well as BiLSTM, and fully connected layer in an end-to-end manner. We adopt this training strategy to train the model with different learning rates for different epochs that helps the model to retrain the pre-trained knowledge of the BERT and avoid catastrophic forgetting of this knowledge during fine-tuning \cite{howard2018, Wang2019}. Detailed training steps are presented in algorithm~\ref{algo2}.

\SetKwInput{KwInput}{Input}                
\SetKwInput{KwOutput}{Output}              

\begin{algorithm}
\DontPrintSemicolon
  
  \KwInput{Training data $D = \left \{ x_n, y_n \right \}_{n=1}^N $}
  
     Load pre-trained BERT model parameters $\theta_1$ \;
     Load pre-trained ConceptNet Embedding \;
     Initialize BiLSTM parameters $\theta_2$ \;
 
   \For {each epoch}{
     Sample mini-batch  $ (x, y) \subset D $ \;
     Construct argument similarity model with Eq~\ref{eq_sim} \;
     Train argument similarity model with a learning rate of $2e-5$ \;
     Update parameters $\theta_1$, $\theta_2$ \;
    }
   \KwOutput{Model parameters}

\caption{Training procedure of argument similarity model}
 \label{algo1} 
\end{algorithm}

\begin{algorithm}[h]
\DontPrintSemicolon
  
  \KwInput{Training data $D = \left \{ x_n, y_n \right \}_{n=1}^N $}
  Load pre-train BERT parameters $\theta_3$ \;
  Initialize BiLSTM parameters $\theta_4$ \;
  Freeze parameters $\theta_3$ \;
   \For {each epoch}{
     Sample mini-batch $ (x, y) \subset D $ \;
     Construct intent classifier model with Eq~\ref{eq_inten_final} \;
     Train intent classifier model with a learning rate of $1e-4$  \;
     Update parameters  $\theta_4$ \;
    }
  Unfreeze parameters $\theta_3$ \;
  \For {each epoch}{
     Sample mini-batch  $ (x, y) \subset D $ \;
     Load trained intent classifier model \;
     Continue train intent classifier model with a learning rate of $2e-5$  \;
     Update parameters  $\theta_3$, $\theta_4$  \;
    }
   \KwOutput{Model parameters}
 

\caption{Training procedure of intent classifier model}
\label{algo2} 
\end{algorithm}


\section{Evaluation and Results}~\label{sec:EvalRes}

\subsection{Evaluation – Intent classification}
We compare the performance of the proposed intent classifier model against the following baseline methods.
\begin{enumerate}
    \item Embedding Classifier: The embedding intent classifier model from Rasa\footnote{https://rasa.com/} NLU inspired by StarSpace \cite{wu2018Star}, counts distinct words of the training data and provides these word token counts as input features to the intent classifier. We trained this model using the Rasa framework for 300 epochs. 
    
    \item Logistic Regression with BERT (LR + BERT): The model extracts features from the pre-trained Bert model. The BERT model processes the user utterances and the final hidden state of the \verb|[CLS]| token of each utterance is passed as features to the logistic regression model. 
    The model then trains logistic regression parameters on these extracted features to predict the user intent. It is noteworthy that parameters of BERT remained fixed during training in this model.    
    
     \item Dual Intent and Entity Transformer (DIET) Classifier:  
     The DIET \cite{bunk2020Diet} model is multi-task architecture for intent classification and entity recognition. The model obtains dense features from pre-trained word embedding models. These features are then used by a 2 layer transformer with relative position attention. The BERT-base model is employed for producing dense features. The model is trained using the Rasa framework for 100 epochs. 
     
      \item BERT Classifier: In the pre-trained BERT model \cite{devlin2019bert}, we add a fully-connected layer on top of the last encoder layer \verb|[CLS]| token for classifying user intent. The model is fine-tuned for 8 epochs with a batch size of 16 and a learning rate of $2e-5$.
     
     \item DistilBERT Classifier: The DistilBERT model \cite{disBERT} utilizes knowledge distillation during pre-training to reduce the size of the BERT model. We added one fully-connected layer on top of the final encoder \verb|[CLS]| token for predicting user intent. The model is fine-tuned using Adam optimizer with a learning rate of $2e-5$ and a batch size of 16 for 8 epochs.
     
     \item RoBERTa Classifier: The RoBERTa model \cite{roberta2018} optimizes the BERT pre-training approach by employing dynamic masking, large mini-batches, and a larger byte-level byte-pair encoding (BPE) for training the robust model. The fully connected layer is applied on top of \verb|[CLS]| token of the final encoder layer for predicting user intent. The model is fine-tuned with Adam optimizer using a learning rate of $2e-5$ and a batch size of 16 for 8 epochs.
     
\end{enumerate}

\begin{table*}
\begin{center}
\caption{Intent classifier performance comparison on Users Study and Banking77 datasets with the different number of training examples i.e., 10-shot (10 training examples per intent), 30-shot (30 training examples per intent), and full training data. For few-shot setups, models were trained with 5 random selection of 10-20-30 examples. Performance is reported in mean accuracy scores $\times$ 100 $\pm$ standard deviation.} \label{tab:intent_results}

\begin{tabular}{l|r|r|r|r|r|r|r|r}
\hline
	& \multicolumn{3}{c}{User Study}  & & \multicolumn{3}{c}{Banking77} \\
\cline{2-9}

Model  & \small{10-shot}& \small{20-shot} & \small{30-shot} & \small{Full} & \small{10-shot}& \small{20-shot} & \small{30-shot} & \small{Full}  \\
 
\hline
Embedding  & 59.6$\pm$1.8 & 65.1$\pm$2.1 & 69.1$\pm$1.3 & 73.1 & 57.9$\pm$1.2 & 71.5$\pm$0.7 & 74.7$\pm$0.4 & 86.2 \\
LR+BERT & 56.4$\pm$1.6 & 67.8$\pm$1.4 &  74.9$\pm$1.2 & 76.3 & 56.1$\pm$1.1 & 68.4$\pm$0.6 & 75.3$\pm$0.3 & 86.9 \\
DIET  & 54.6$\pm$2.5 & 68.8$\pm$2.6 & 75.8$\pm$1.7 & 82.3 & 55.6$\pm$1.3 & 76.3$\pm$0.8 & 82.6$\pm$0.3 & 90.3\\
DistilBERT & 68.3$\pm$2.2 & 79.6$\pm$2.7 & 83.5$\pm$2.4 & 88.2 & 76.2$\pm$1.8 & 84.6$\pm$0.5 & 87.6$\pm$0.3 & 92.9 \\
BERT Classifier  & 70.5$\pm$2.4 & 81.8$\pm$1.8 & 84.8$\pm$2.9 & 89.3 & 80.1$\pm$2.1 & 86.3$\pm$0.7 & 88.5$\pm$0.3 & 93.2  \\
RoBERTa Classifier & 65.1$\pm$3.5 & 80.2$\pm$1.2 & 84.0$\pm$2.5 & 89.3 & 74.3$\pm$2.2 & 86.1$\pm$1.4 & 88.4$\pm$0.9 & 93.2  \\
BERT+BiLSTM +ArgSim & \textbf{73.4$\pm$2.3} &  \textbf{84.6$\pm$1.7} &  \textbf{86.6$\pm$2.6} & \textbf{89.7} & \textbf{81.8$\pm$1.9} &  \textbf{87.4$\pm$0.8}& \textbf{90.1$\pm$0.3} & \textbf{93.9} \\
\hline
\end{tabular}
\end{center}
\end{table*}


Intent classification results are presented in Table \ref{tab:intent_results} (statistically significant with $p < 0.05$ under t-test). The results demonstrate that the embedding classifier performs poorly in 30-shot and full data against all baseline approaches because it utilizes merely the word counts and does not consider pre-trained language model information. In contrast, LR + BERT model obtained better results than the embedding classifier, because it extracts features from the pre-trained language model for predicting user intent. Furthermore, the DIET classifier employed 2 Transformer encoder layers to learn the contextualized sentence representation and outperformed the former models. The only case where the DIET classifier performs poorly than the embedding classifier is the 10-shot case. The reason behind this is the lack of training data as the pre-trained language model employed by the DIET model requires more training data to generalize well. Besides, DistilBERT, BERT, and RoBERTa models achieved superior results than the DIET classifier, because these models employ 6, 12, and 12 Transformer encoder layers respectively, as compared to 2 Transformer encoder layers used by the DIET classifier, therefore, these models provide better and robust utterance representation. 
We observe the marginal differences in the performance of the BERT and RoBERTa classifiers. Furthermore, the RoBERTa classifier performs better than DistilBERT in most cases. The only setup where  DistilBERT performs better than the RoBERTa is a 10-shot setup. This indicates that RoBERTa requires a large amount of data to generalize well. 
Nevertheless, our proposed model outperforms all baseline methods and performance increases by approximately 16\%, 13\%, and 7\% better accuracy score compared to embedding classifier, LR + BERT, and DIET classifier, respectively on the full data setup of the user study dataset. Furthermore, the improvement over embedding classifier, LR + BERT, and DIET classifier are 7.7\%, 7\%, and 3.7\% on the full data setup of Banking77 dataset. This is because the proposed BERT+BiLSTM+ArgSim model fine-tuned the BERT model to obtain contextual utterance representation for predicting user intent. Additionally, we observe that stacking BiLSTM on top of the BERT model and concatenating sentence representation from the argument similarity model provides better results than stacking just a single fully-connected layer.    

\begin{figure}[]
  \centering
  \begin{minipage}[b]{0.45\textwidth}
    \includegraphics[width=\textwidth]{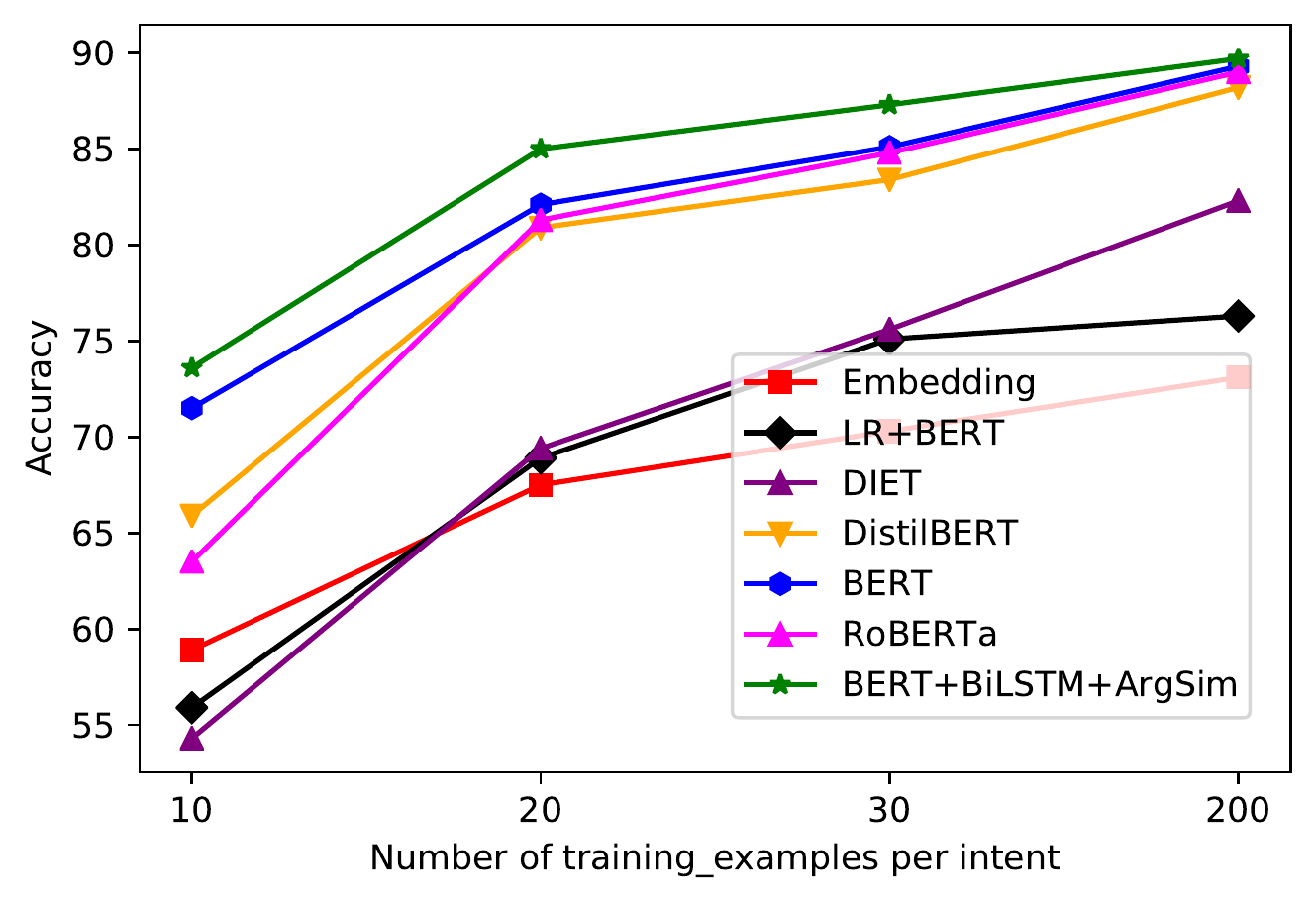}
    \caption{Performance comparison of the intent classifier on the User study dataset with respect to the number of training examples per intent.}
  \label{fig_few_user}
  \end{minipage}
  \hfill
  \begin{minipage}[b]{0.45\textwidth}
    \includegraphics[width=\textwidth]{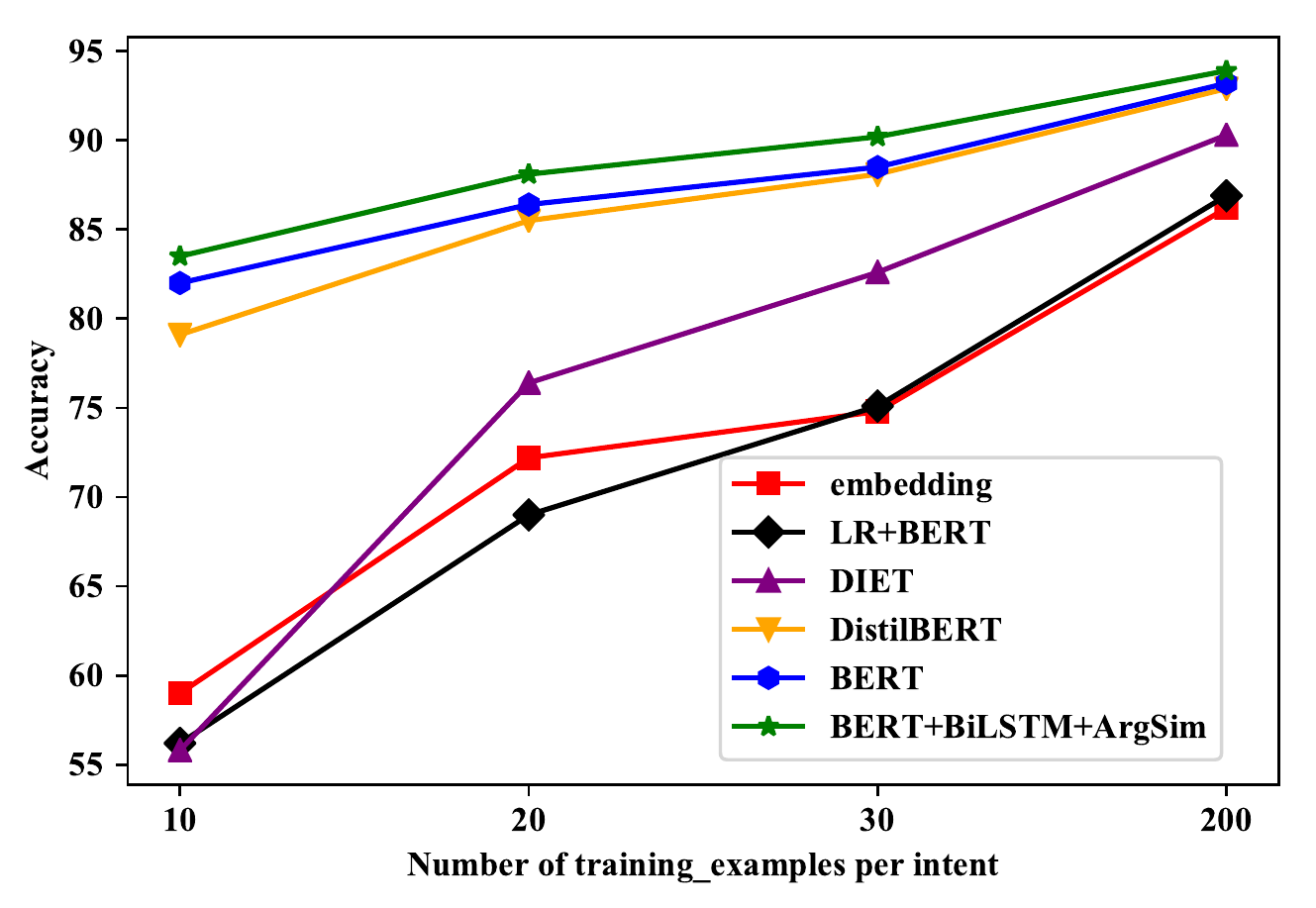}
    \caption{Performance comparison of the intent classifier on the Banking77 dataset with respect to the number of training examples per intent.}
    \label{fig_few_bank}
  \end{minipage}
\end{figure}

From Fig.~\ref{fig_few_user} and Fig.~\ref{fig_few_bank}, we can observe that in the few-shot scenario i.e., 10-shot, 20-shot, and 30-shot; our  BERT + BiLSTM+ArgSim model performance gains are more prominent over baselines models. The improvement gains of our model over the state-of-the-arts DistilBERT, RoBERTa, and BERT classifier models are almost 5\%, 8\%, and 3\% for the user study dataset and 6\%, 7.5\%, and 1.7\% for the Banking77 dataset in a 10-shot setup (10 samples per intent). Similarly, we see the proposed model outperforms all model in the 20-shot and 30-shot setup on both datasets. Overall, our model has a clear advantage over other methods in all setups, and more prominently in few-shot setups. Furthermore, the proposed model is statistically significant as compared to Embedding, LR+BERT, DIET, DistilBERT, RoBERTa, and BERT models with $p$-value of $1e-05$, $2e-05$, $ 0.0019$, $0.023$, $0.044$, and $0.049$ respectively, on user study dataset.

\subsection{Evaluation – Argument Similarity}

We evaluate the performance of the proposed argument similarity model for identifying the reference argument task and also validate our model performance for the common semantic textual similarity task. We perform experiments on user study and semantic textual similarity benchmark dataset. We compared the results against the following baseline methods.

\begin{enumerate} 
    \item Average of glove embeddings;
    \item Average of ConceptNet Numberbatch embeddings;
    \item Mean of last layer tokens representations of BERT;
    \item InferSent model~\cite{conneau2017InferSent} is trained on a natural language inference dataset using the siamese BiLSTM network structure with max-pooling over the output.
    \item Universal Sentence Encoder~\cite{cer2018universal} is a strong sentence-level embedding model trained using transformer architecture and multiple objectives.
    
    \item Sentence-BERT model~\cite{Reimers2019SBERT} a state-of-the-art sentence embedding model is trained using a siamese network structure over BERT. 
    We fine-tune SBERT base and large models with a learning rate of $2e-5$ using Adam optimizer on the STSB dataset. The mean pooling strategy is employed to mean output vectors of the last layer of BERT for producing a fixed-sized sentence embedding.

    \item Sentence-RoBERTa model~\cite{Reimers2019SBERT} SRoBERTa fine-tunes the RoBERTa pre-trained model to produce robust sentence embeddings which can be compared using cosine similarity. We fine-tune the RoBERTa network on the STSB dataset with Adam optimizer using a learning rate of $2e-5$. We report experiments using the mean pooling strategy.

    \item Sentence-XLM-Align model fine-tunes cross-lingual language models word alignment (XLM-Align) pre-trained model~\cite{chi-improving} to generate useful sentence embeddings. We fine-tune the pre-trained XLM-Align model on the STSB dataset using Adam optimizer with a learning rate of $2e-5$. The mean pooling strategy is applied to produce fixed-sized sentence embeddings.

\end{enumerate} 

\begin{table}
\begin{center}
\caption{Argument similarity model performance comparison on User Study and STS datasets. SBERT-STSb-base, SBERT-STSb-large, and ArgSim models are trained on the STS-B dataset. Performance on user study is reported in the accuracy scores $\times$ 100. and performance on STS is reported in SPEARMAN $\times$ 100.} \label{tab:argSim_results}
\begin{tabular}{l|c|c}
\hline
Model &  User Study & STSB \\
      & Accuracy    & SPEARMAN  \\
\hline
Avg. GloVe  & 93.2 & 61.5 \\
Avg. ConceptNet  &  94.0 & 65.1 \\
Avg. BERT  & 93.2 & 47.2\\
InferSent - Glove  & 90.2 & 75.8 \\
USE &   \textbf{95.2} & 78.2 \\
SBERT-STSb-base  & 94.0 & 84.6\\
SBERT-STSb-large & 94.0 & 84.4 \\
SRoBERTa-STSb-base  & 94.4 & 84.8 \\
SXLM-Align-STSb-base & 93.8 & 80.5 \\
ArgSim (Ours work) &   \textbf{95.2} & \textbf{85.1} \\
\hline
\end{tabular}

\end{center}
\end{table}


Table~\ref{tab:argSim_results}  presents the results corresponding to the argument similarity task. The results generated by our model outperform InferSent, SXLM-Align, and SBERT by achieving approximately 4\%, 2\%, and 2\% better accuracy on the user study dataset, respectively. The SXLM-Align model produces worse results than SBERT and SRoBERTa on both the user study dataset and STSB dataset. On the other hand, the SRoBERTa model performs slightly better than the SBERT model on both datasets. 
Also, USE performs better than InferSent and SBERT on the user study dataset, as it is pre-trained on question answering data in addition to NLI data, which is related to the classification task. Our model performance matches the performance of the state-of-the-art USE model on the user study dataset. 
Furthermore, the mean of last layer tokens representations of BERT embeddings, average glove embeddings, and an average of ConceptNetNumberbatch embeddings perform poorly on the STSB dataset. However, for the user study dataset, these methods produced better results than the supervised InferSent model trained using the siamese structure on NLI data. The reason behind this is that for identifying the reference argument task, we calculate the cosine-similarity between candidate arguments and current utterance, and the closest argument to the current utterance embedding is selected as a reference argument. This allows two-sentence embeddings to have high and low similarity on certain dimensions and still correctly identify a reference argument. In contrast, the STS task is a regression task, which estimates the similarities between two-sentence embeddings by cosine-similarity and treats all dimensions equally. This indicates average word embeddings are infeasible for the STS task.  
Nonetheless, our model trained on the STS benchmark yields better sentence representation for argument similarity task and STS task as it combines contextualized word representation with common sense knowledge obtained from ConceptNet Numberbatch embeddings.

\subsection{Evaluation- Complete Pipeline}
We evaluate the performance of the complete pipeline consisting of intent classification and argument similarity modules for native English speakers and non-native English speakers. The performance of intent classification and argument similarity module is measured in F1 score and accuracy, respectively. The complete pipeline performance is measured in an accuracy matrix. The overall accuracy of the complete pipeline is the percentage of utterances where the pipeline correctly predicts both intents and presented arguments. The model is trained on the native speakers dataset and evaluated on native and non-native speakers test-set. The statistics of train and test samples are given in table~\ref{tab:class_dist}.  The results are shown in table~\ref{tab:pipeline_results}. We can observe from the results that both intent classifier and argument similarity models perform better on the native speaker dataset. The respective improvements are around 1\% and 5\% for intent classifier and argument similarity models. However, we do not observe a significant difference in accuracy between native speakers and non-native speakers on the complete pipeline. Overall, this proves that the proposed framework is robust to the different language proficiency of the users.

\begin{table}
\begin{center}
\caption{
Complete pipeline performance of the proposed framework on native speakers and Non-native speakers datasets.} \label{tab:pipeline_results}
\begin{tabular}{l|r|r|r}
\hline
Model &  Intent F1 & Sim Acc. &  Overall Acc.\\
\hline
Native Speakers &  89.8 &  95.2  & 87.7\\
Non-native Speakers &  88.8 &  89.7  & 87.3 \\
\hline
\end{tabular}

\end{center}
\end{table}

\subsection{Impact of hyper-parameters}
\begin{figure}[]
  \centering
  \begin{minipage}[b]{0.45\textwidth}
    \includegraphics[width=\textwidth]{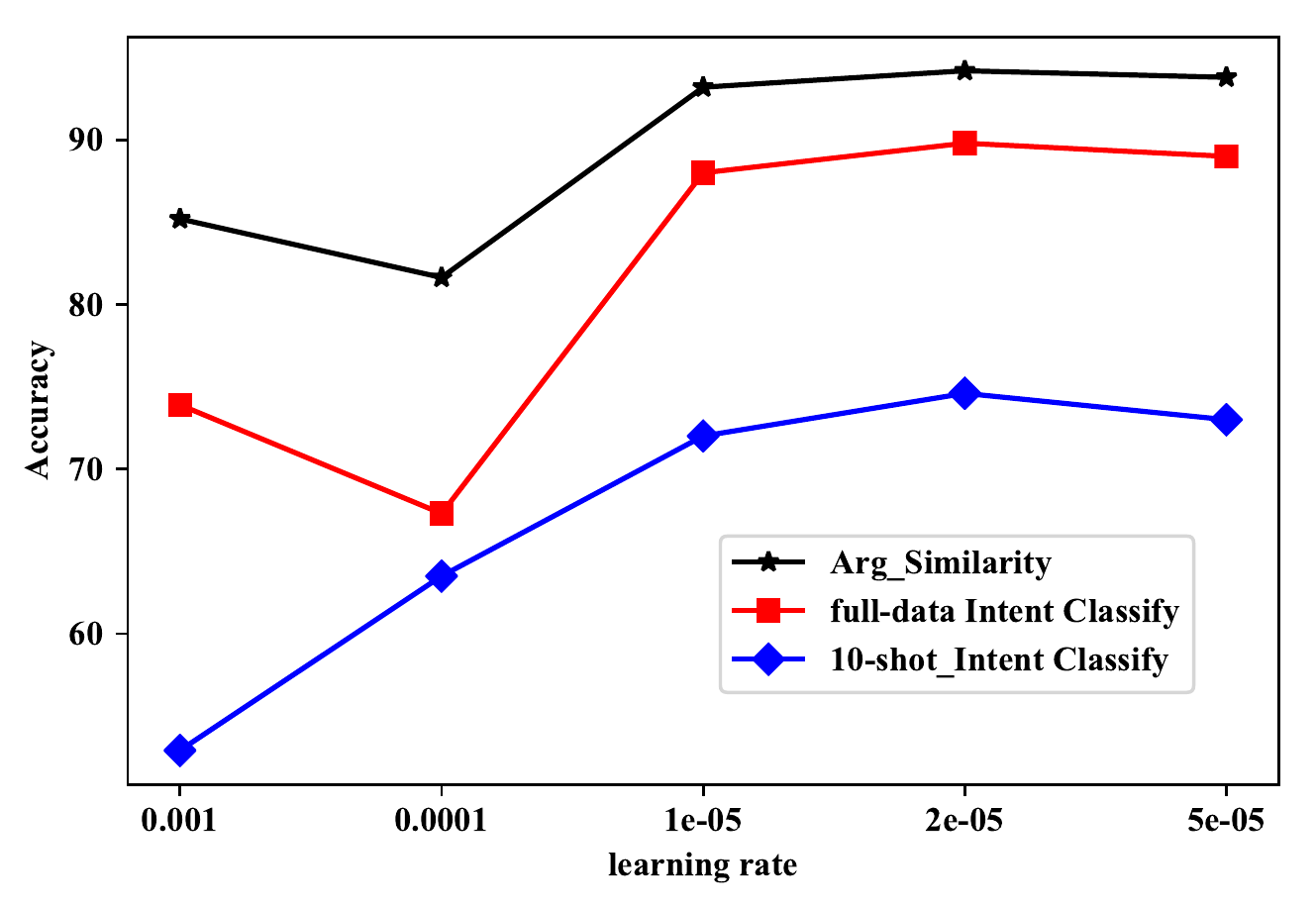}
    \caption{Effect of learning rate on model performance for intent classification and argument similarity tasks on User study data set.}
  \label{fig_lr_user}
  \end{minipage}
  \hfill
  \begin{minipage}[b]{0.45\textwidth}
    \includegraphics[width=\textwidth]{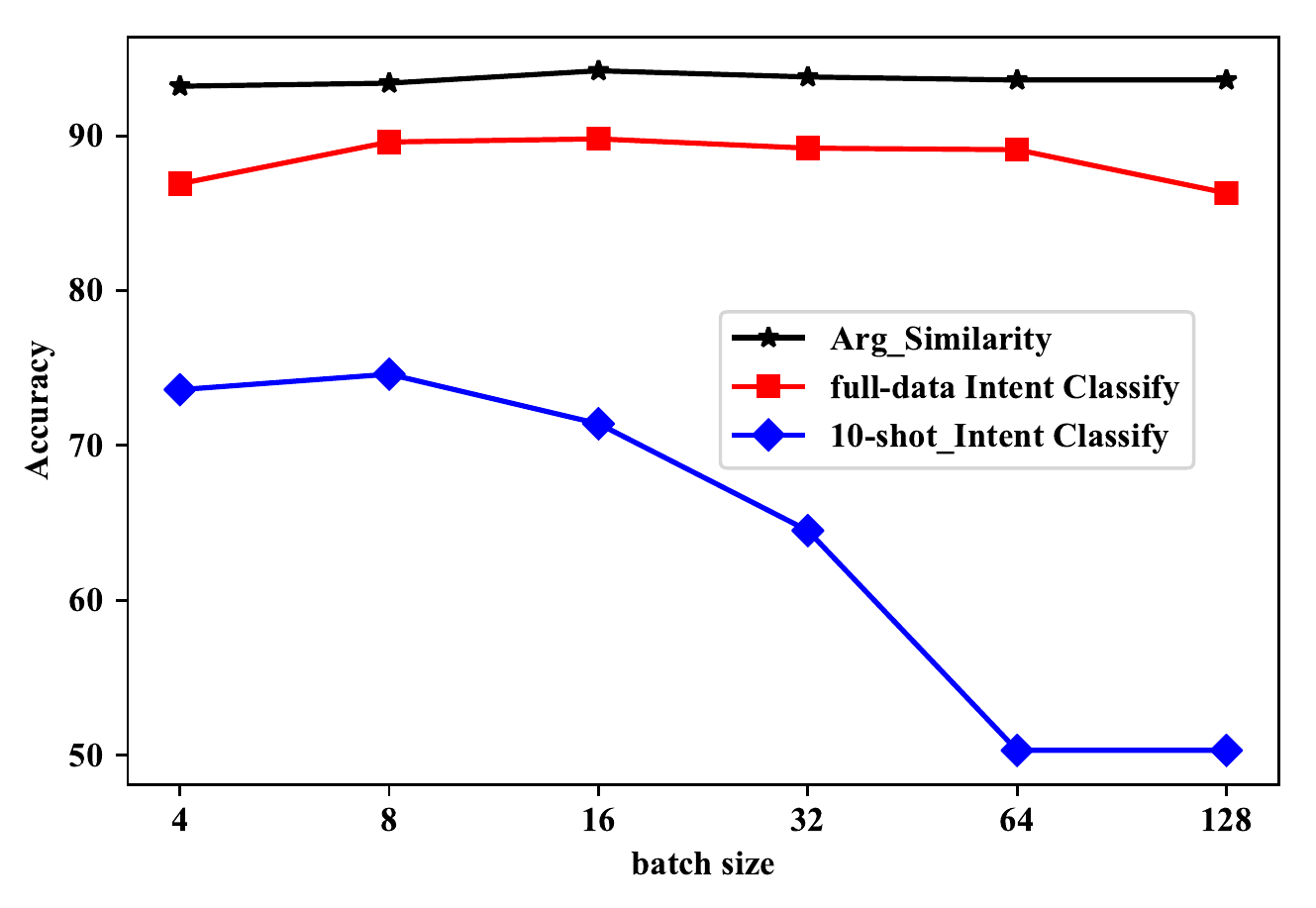}
    \caption{Effect of batch size on model performance for intent classification and argument similarity tasks on User study data set.}
    \label{fig_batch_user}
  \end{minipage}
  
\end{figure}

\begin{figure}[]
  \centering
  \begin{minipage}[b]{0.45\textwidth}
    \includegraphics[width=\textwidth]{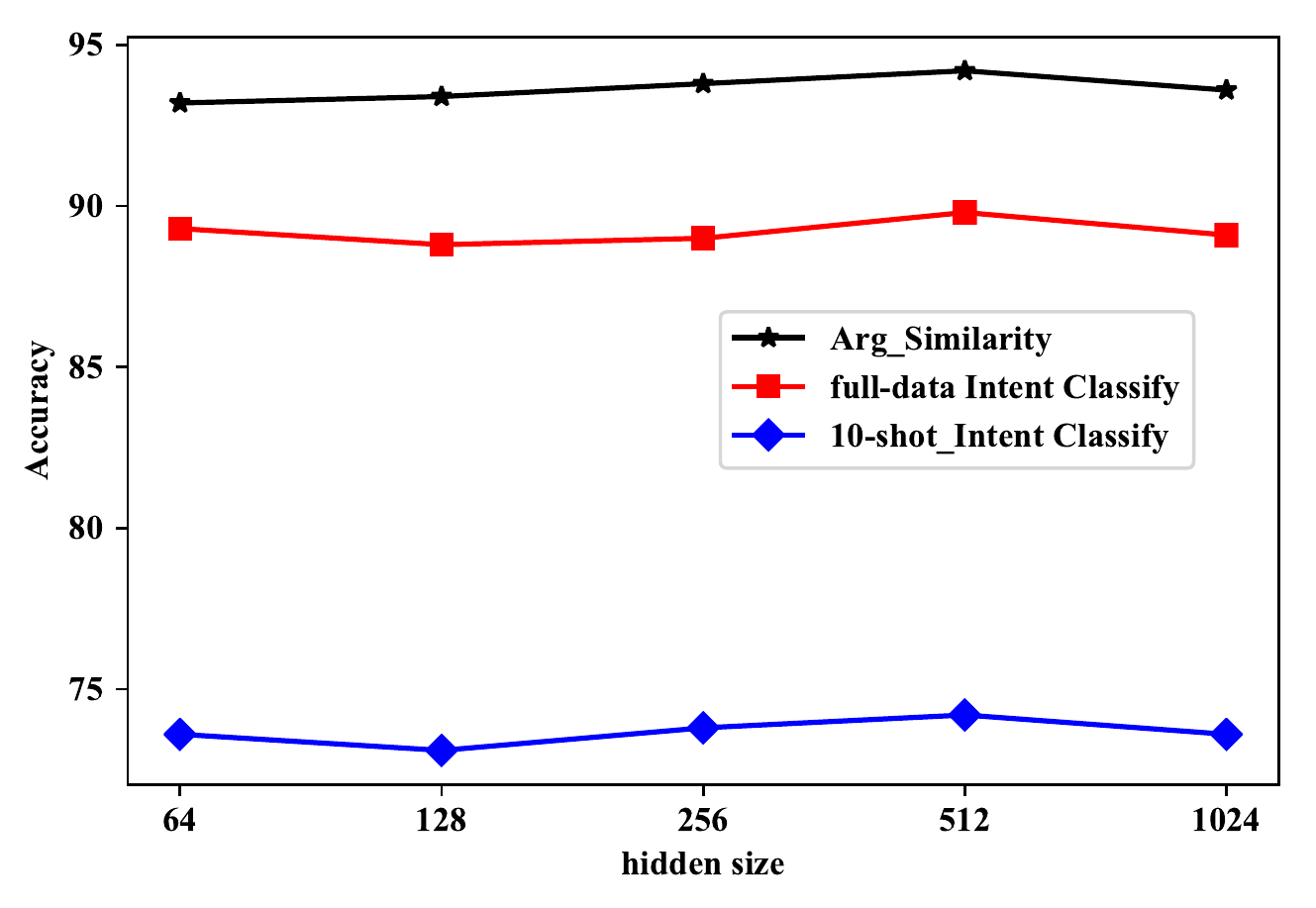}
    \caption{Effect of hidden size on model performance for intent classification and argument similarity tasks on User study data set.}
  \label{fig_hidden_size}
  \end{minipage}
  \hfill
  
\end{figure}

The learning rate, batch size, and hidden size are the most important hyper-parameters of the proposed model. The model performance is analyzed concerning these parameters. We employ the random search~\cite{bergstra2012random} strategy for finding hyper-parameters randomly from a specified subset of hyper-parameters. We choose the learning rate from  $ \lbrack 0.001, 0.0001, 1e-5, 2e-5, 5e-5 \rbrack$. In Fig.~\ref{fig_lr_user} we can see that the learning rate has a significant impact on model performance for both intent classification and argument similarity tasks. Furthermore, we observe that the proposed model performs poorly on learning rates of $0.001$ and $0.0001$, which shows that the model fails to converge on high learning rates. In most cases, a lower learning rate produces better results. Especially when the learning rate is $2e-05$ model achieves the highest accuracy on the user study dataset for both tasks. Furthermore, we choose a batch size from a range of $ \lbrack 4, 8, 16, 32, 64, 128 \rbrack$. Fig. \ref{fig_batch_user} reveals that change in batch size has less impact on model performance for argument similarity task and full data setup of intent classification. The model produces a higher accuracy on batch sizes of 16 and 32 for the full data setup of intent classification. However, batch size has a significant impact on model performance for the 10-shot intent classification setup. The model performance increases with a small batch size of $4$ and $8$. Especially when batch size is $8$, the model achieves the highest accuracy of $74.6\%$. As the value of batch size increases, the number of mini-batches decreases, and model performance decreases. 
We explore the following values $ h = \lbrack 64, 128, 256, 512, 1024 \rbrack$ for hidden dimensions of LSTM. From Fig.~\ref{fig_hidden_size} we can observe that the performance of the model does not change significantly for different sizes of hidden dimensions of LSTM. Results suggest that the proposed model is robust to hidden dimensions of LSTM for intent classification and argument similarity tasks. Furthermore, a lower learning rate and smaller batch size yield better results especially for the 10-shot  intent classification setup.

\begin{table*}
\begin{center}
\caption{Evaluation results of intent classifier ablation on the Users Study and Banking77 datasets for 10-shot, 20-shot, 30-shot, and full training data setups.} \label{tab:intent_abl}
\begin{tabular}{l|r|r|r|r|r|r|r|r}
\hline
 & \multicolumn{3}{c}{User Study}  &  & \multicolumn{3}{c}{Banking77} \\
\cline{2-9}

Model  & 10-shot & 20-shot & 30-shot & Full-data & 10-shot & 20-shot & 30-shot & Full-data  \\
 
\hline

BERT-tuned & 71.5 & 82.1 & 85.1 & 89.3                 & 82.0 & 86.7 & 88.5 & 93.2  \\
BERT+BiLSTM  & 72.8 & 83.2  & 85.7 & 89.5           & 82.6  & 87.4 & 89.2 & 93.4  \\
BERT+BiLSTM
+ ArgSim & \textbf{74.6} & \textbf{85.0} & \textbf{87.3} & \textbf{89.7} & \textbf{83.5} & \textbf{88.1}& \textbf{90.2} & \textbf{93.9} \\
\hline

\end{tabular}

\end{center}
\end{table*}

\subsection{Ablation Study}

To demonstrate the effectiveness of different aspects of the intent classifier model, we conduct an ablation study on the two datasets. The results are shown in table~\ref{tab:intent_abl}. It shows that adding BiLSTM and ArgSim representation improves the performance of the plain fine-tuned BERT model. Adding BiLSTM provides performance improvement of 1.3\%, 1.1\%, 0.6\%, and 0.2\% for 10-shot, 20-shot, 30-shot, and full-data respectively on user study dataset, and 0.6\%,  0.7\%, 0.7\%, and 0.2\% improvement on the  Banking77 dataset. Furthermore, adding representation from the argument similarity model further improves the performance of the proposed model especially in the 10-shot, 20-shot, and 30-shot scenario and corresponding improvements on user study dataset are 1.8\%, 1.8\%, 1.6\% respectively. The overall performance gains over BERT-tuned are 3.1\%, 2.9\%, 2.2\%, and 0.4\% respectively for the user study dataset, and 1.5\%, 1.4\%, 1.2\% and, 0.7\% improvement for the Banking77 dataset. 

We start the ablation of argument similarity (ArgSim) model by removing different components in our model to get an understanding of their importance. We first remove the BERT part (-BERT) to check model performance with only the BiLSTM part, and then remove the BiLSTM part (-BiLSTM) to check model performance with only the BERT part. Table~\ref{tab:argSim_abl} shows the ablation results on user study and STS dataset. From the results, we observe that the BERT part plays a significant role in model performance, and removing the BERT part decreases the model performance by around 2\% in terms of accuracy on user study dataset and 14.1\% in terms of spearman correlation on STSB dataset. Next, we show that the BiLSTM part is also an important contributor to model performance, and removing the BiLSTM part decreases the model performance by around 1.2\% in terms of accuracy on user study dataset and 1.1\% in terms of spearman correlation on STSB dataset. These results indicate both BERT and LSTM parts are important for the argument similarity model.

\begin{table}
\begin{center}
\caption{Evaluation results of argument similarity model ablation on User Study and STSB datasets.} \label{tab:argSim_abl}
\begin{tabular}{l|r|r}
\hline
Model &  User Study & STSB \\
      & Accuracy    & SPEARMAN  \\
\hline
ArgSim  &   \textbf{95.2} & \textbf{85.1} \\
- BiLSTM  & 94.0 & 84.0 \\
- BERT  & 93.2  & 71.0 \\

\hline
\end{tabular}
\end{center}
\end{table}

\section{Discussion and Conclusion}~\label{sec:Conc}
Throughout this work, we introduced an NLU approach for argumentative dialogue systems in the domain of information seeking and opinion building. Our approach detects arguments addressed by the user within his or her utterance and distinguishes between multiple intents including user preferences towards the respective arguments. Our approach was applied and tested in an actual argumentative dialogue system on data collected in an extensive user study. Additionally, we evaluated the proposed intent classifier and argument similarity models on the Banking77 and STS benchmark datasets.
 Throughout the evaluation, we assessed the performance of the NLU components against state-of-the-art baselines on different datasets, the robustness of the proposed approach against new topics, and the robustness of the approach against different language proficiency and cultural diversity. Besides, the performance of the intent classifier model is assessed in full data and few-shot setups. Our results show a clear advantage of our model against baselines approaches for intent classification in both full data and few-shot setups. Furthermore, results show the superior accuracy of the proposed model against baselines models for argument similarity tasks as well as the accuracy of 87.7\% in complete pipeline testing. Moreover, no significant difference between utterances from UK users and Chinese users was detected. The results indicate that our model has to be trained only once for each system in order to learn the required system-specific intents but does not require pre-training for new topics or user groups which ensures high flexibility of the respective system.  

Our future work will be focused on two separate aspects. First, we want to explore different confirmation strategies in the dialogue management in order to further improve the recognition rate of the complete system. Moreover, we will extend the capacities of the NLU to allow the user to introduce new arguments which will be learned by the system during the discussion.

\section*{Acknowledgements}

This work has been partially funded by the Deutsche Forschungsgemeinschaft (DFG)
within the project "How to Win Arguments - Empowering Virtual Agents to
Improve their Persuasiveness", Grant Number 376696351, as part of the
Priority Program "Robust Argumentation Machines (RATIO)" (SPP-1999) and the Natural Science Foundation of China grant (U21A20488).

\bibliographystyle{unsrt}  
\bibliography{template}  

\begin{thebibliography}{10}

\bibitem{Lawrence2020}
John Lawrence and Chris Reed.
\newblock Argument mining: A survey.
\newblock {\em Computational Linguistics}, 45(4):765--818, 2020.

\bibitem{niklas2018EVA}
Niklas Rach, Klaus Weber, Louisa Pragst, Elisabeth Andr{\'e}, Wolfgang Minker,
  and Stefan Ultes.
\newblock Eva: a multimodal argumentative dialogue system.
\newblock In {\em Proceedings of the 20th ACM International Conference on
  Multimodal Interaction}, pages 551--552, 2018.

\bibitem{Sakai2020}
Kazuki Sakai, Ryuichiro Higashinaka, Yuichiro Yoshikawa, Hiroshi Ishiguro, and
  Junji Tomita.
\newblock Hierarchical argumentation structure for persuasive argumentative
  dialogue generation.
\newblock {\em IEICE TRANSACTIONS on Information and Systems}, 103(2):424--434,
  2020.

\bibitem{hunter2019}
Anthony Hunter, Lisa Chalaguine, Tomasz Czernuszenko, Emmanuel Hadoux, and
  Sylwia Polberg.
\newblock Towards computational persuasion via natural language argumentation
  dialogues.
\newblock In {\em Joint German/Austrian Conference on Artificial Intelligence
  (K{\"u}nstliche Intelligenz)}, pages 18--33. Springer, 2019.

\bibitem{shigehalli2020nlu}
Pavan~Rajashekhar Shigehalli.
\newblock Natural language understanding in argumentative dialogue systems.
\newblock 2020.

\bibitem{devlin2019bert}
Jacob Devlin, Ming-Wei Chang, Kenton Lee, and Kristina Toutanova.
\newblock {BERT}: Pre-training of deep bidirectional transformers for language
  understanding.
\newblock In {\em Proceedings of the 2019 Conference of the North {A}merican
  Chapter of the Association for Computational Linguistics: Human Language
  Technologies, Volume 1 (Long and Short Papers)}, pages 4171--4186,
  Minneapolis, Minnesota, June 2019. Association for Computational Linguistics.

\bibitem{liu2019linguistic}
Nelson~F. Liu, Matt Gardner, Yonatan Belinkov, Matthew~E. Peters, and Noah~A.
  Smith.
\newblock Linguistic knowledge and transferability of contextual
  representations.
\newblock In {\em Proceedings of the 2019 Conference of the North {A}merican
  Chapter of the Association for Computational Linguistics: Human Language
  Technologies, Volume 1 (Long and Short Papers)}, pages 1073--1094,
  Minneapolis, Minnesota, June 2019. Association for Computational Linguistics.

\bibitem{Cer2017STS}
Daniel Cer, Mona Diab, Eneko Agirre, I{\~n}igo Lopez-Gazpio, and Lucia Specia.
\newblock {S}em{E}val-2017 task 1: Semantic textual similarity multilingual and
  crosslingual focused evaluation.
\newblock In {\em Proceedings of the 11th International Workshop on Semantic
  Evaluation ({S}em{E}val-2017)}, pages 1--14, Vancouver, Canada, August 2017.
  Association for Computational Linguistics.

\bibitem{speer2017conceptnet}
Robyn Speer, Joshua Chin, and Catherine Havasi.
\newblock Conceptnet 5.5: An open multilingual graph of general knowledge.
\newblock In {\em Thirty-First AAAI Conference on Artificial Intelligence},
  pages 4444--4451, 2017.

\bibitem{bea2019}
Annalena Aicher, Niklas Rach, Wolfgang Minker, and Stefan Ultes.
\newblock Opinion building based on the argumentative dialogue system bea.
\newblock In {\em Proceedings of the 10th International Workshop on Spoken
  Dialog Systems Technology (IWSDS 2019)}, 2019.

\bibitem{kim2014cnn}
Yoon Kim.
\newblock Convolutional neural networks for sentence classification.
\newblock In {\em Proceedings of the 2014 Conference on Empirical Methods in
  Natural Language Processing ({EMNLP})}, pages 1746--1751, Doha, Qatar,
  October 2014. Association for Computational Linguistics.

\bibitem{Zhang2015}
Xiang Zhang, Junbo Zhao, and Yann LeCun.
\newblock Character-level convolutional networks for text classification.
\newblock In {\em Advances in neural information processing systems}, pages
  649--657, 2015.

\bibitem{Ravuri2015}
Suman Ravuri and Andreas Stolcke.
\newblock Recurrent neural network and lstm models for lexical utterance
  classification.
\newblock In {\em Sixteenth Annual Conference of the International Speech
  Communication Association}, 2015.

\bibitem{Abro2019Intent}
Waheed~Ahmed Abro, Guilin Qi, Huan Gao, Muhammad~Asif Khan, and Zafar Ali.
\newblock Multi-turn intent determination for goal-oriented dialogue systems.
\newblock In {\em 2019 International Joint Conference on Neural Networks
  (IJCNN)}, pages 1--8. IEEE, 2019.

\bibitem{liu2016attention}
Bing Liu and Ian Lane.
\newblock Attention-based recurrent neural network models for joint intent
  detection and slot filling.
\newblock In {\em INTERSPEECH 2016}, pages 685--689, 2016.

\bibitem{ABRO2020KBS}
Waheed~Ahmed Abro, Guilin Qi, Zafar Ali, Yansong Feng, and Muhammad Aamir.
\newblock Multi-turn intent determination and slot filling with neural networks
  and regular expressions.
\newblock {\em Knowledge-Based Systems}, page 106428, 2020.

\bibitem{goo2018slot}
Chih-Wen Goo, Guang Gao, Yun-Kai Hsu, Chih-Li Huo, Tsung-Chieh Chen, Keng-Wei
  Hsu, and Yun-Nung Chen.
\newblock Slot-gated modeling for joint slot filling and intent prediction.
\newblock In {\em Proceedings of the 2018 Conference of the North {A}merican
  Chapter of the Association for Computational Linguistics: Human Language
  Technologies, Volume 2 (Short Papers)}, pages 753--757, New Orleans,
  Louisiana, June 2018. Association for Computational Linguistics.

\bibitem{bunk2020Diet}
Tanja Bunk, Daksh Varshneya, Vladimir Vlasov, and Alan Nichol.
\newblock {DIET:} lightweight language understanding for dialogue systems.
\newblock {\em CoRR}, abs/2004.09936, 2020.

\bibitem{convert-2020}
Matthew Henderson, I{\~n}igo Casanueva, Nikola Mrk{\v{s}}i{\'c}, Pei-Hao Su,
  Tsung-Hsien Wen, and Ivan Vuli{\'c}.
\newblock {C}onve{RT}: Efficient and accurate conversational representations
  from transformers.
\newblock In {\em Findings of the Association for Computational Linguistics:
  EMNLP 2020}, pages 2161--2174, Online, November 2020. Association for
  Computational Linguistics.

\bibitem{dual-encoder-2020}
I{\~n}igo Casanueva, Tadas Tem{\v{c}}inas, Daniela Gerz, Matthew Henderson, and
  Ivan Vuli{\'c}.
\newblock Efficient intent detection with dual sentence encoders.
\newblock In {\em Proceedings of the 2nd Workshop on Natural Language
  Processing for Conversational AI}, pages 38--45, Online, July 2020.
  Association for Computational Linguistics.

\bibitem{cer2018universal}
Daniel Cer, Yinfei Yang, Sheng-yi Kong, Nan Hua, Nicole Limtiaco, Rhomni
  St.~John, Noah Constant, Mario Guajardo-Cespedes, Steve Yuan, Chris Tar,
  Brian Strope, and Ray Kurzweil.
\newblock Universal sentence encoder for {E}nglish.
\newblock In {\em Proceedings of the 2018 Conference on Empirical Methods in
  Natural Language Processing: System Demonstrations}, pages 169--174,
  Brussels, Belgium, November 2018. Association for Computational Linguistics.

\bibitem{Minaee2021}
Shervin Minaee, Nal Kalchbrenner, Erik Cambria, Narjes Nikzad, Meysam
  Chenaghlu, and Jianfeng Gao.
\newblock Deep learning--based text classification: A comprehensive review.
\newblock {\em ACM Computing Surveys}, 54(3), 2021.

\bibitem{ali2020graph}
Zafar Ali, Guilin Qi, Pavlos Kefalas, Waheed~Ahmad Abro, and Bahadar Ali.
\newblock A graph-based taxonomy of citation recommendation models.
\newblock {\em Artificial Intelligence Review}, 53(7), 2020.

\bibitem{ali2020paper}
Zafar Ali, Guilin Qi, Khan Muhammad, Bahadar Ali, and Waheed~Ahmed Abro.
\newblock Paper recommendation based on heterogeneous network embedding.
\newblock {\em Knowledge-Based Systems}, 210:106438, 2020.

\bibitem{Akhtar2020intense}
Md~Shad Akhtar, Asif Ekbal, and Erik Cambria.
\newblock How intense are you? predicting intensities of emotions and
  sentiments using stacked ensemble.
\newblock {\em IEEE Computational Intelligence Magazine}, 15(1):64--75, 2020.

\bibitem{BASIRI2021279}
Mohammad~Ehsan Basiri, Shahla Nemati, Moloud Abdar, Erik Cambria, and
  U.~Rajendra Acharya.
\newblock Abcdm: An attention-based bidirectional cnn-rnn deep model for
  sentiment analysis.
\newblock {\em Future Generation Computer Systems}, 115:279--294, 2021.

\bibitem{Cambria2020SenticNet}
Erik Cambria, Yang Li, Frank~Z. Xing, Soujanya Poria, and Kenneth Kwok.
\newblock {\em SenticNet 6: Ensemble Application of Symbolic and Subsymbolic AI
  for Sentiment Analysis}, page 105–114.
\newblock Association for Computing Machinery, New York, NY, USA, 2020.

\bibitem{Skip-Thought2015}
Ryan Kiros, Yukun Zhu, Russ~R Salakhutdinov, Richard Zemel, Raquel Urtasun,
  Antonio Torralba, and Sanja Fidler.
\newblock Skip-thought vectors.
\newblock In C.~Cortes, N.~Lawrence, D.~Lee, M.~Sugiyama, and R.~Garnett,
  editors, {\em Advances in Neural Information Processing Systems}, volume~28,
  pages 3294--3302. Curran Associates, Inc., 2015.

\bibitem{word2vec2013}
Tomas Mikolov, Ilya Sutskever, Kai Chen, Greg~S Corrado, and Jeff Dean.
\newblock Distributed representations of words and phrases and their
  compositionality.
\newblock In C.~J.~C. Burges, L.~Bottou, M.~Welling, Z.~Ghahramani, and K.~Q.
  Weinberger, editors, {\em Advances in Neural Information Processing Systems},
  volume~26, pages 3111--3119. Curran Associates, Inc., 2013.

\bibitem{hill-2016-learning}
Felix Hill, Kyunghyun Cho, and Anna Korhonen.
\newblock Learning distributed representations of sentences from unlabelled
  data.
\newblock In {\em Proceedings of the 2016 Conference of the North {A}merican
  Chapter of the Association for Computational Linguistics: Human Language
  Technologies}, pages 1367--1377, San Diego, California, June 2016.
  Association for Computational Linguistics.

\bibitem{conneau2017InferSent}
Alexis Conneau, Douwe Kiela, Holger Schwenk, Lo\"{i}c Barrault, and Antoine
  Bordes.
\newblock Supervised learning of universal sentence representations from
  natural language inference data.
\newblock In {\em Proceedings of the 2017 Conference on Empirical Methods in
  Natural Language Processing}, pages 670--680, Copenhagen, Denmark, September
  2017. Association for Computational Linguistics.

\bibitem{Peters2018}
Matthew~E. Peters, Mark Neumann, Mohit Iyyer, Matt Gardner, Christopher Clark,
  Kenton Lee, and Luke Zettlemoyer.
\newblock Deep contextualized word representations.
\newblock In {\em Proc. of NAACL}, 2018.

\bibitem{radford2018improving}
Alec Radford, Karthik Narasimhan, Tim Salimans, and Ilya Sutskever.
\newblock Improving language understanding by generative pre-training (2018).
\newblock {\em URL https://s3-us-west-2. amazonaws.
  com/openai-assets/research-covers/language-unsupervised/language\_
  understanding\_paper. pdf}, 2018.

\bibitem{yang2019xlnet}
Zhilin Yang, Zihang Dai, Yiming Yang, Jaime Carbonell, Russ~R Salakhutdinov,
  and Quoc~V Le.
\newblock Xlnet: Generalized autoregressive pretraining for language
  understanding.
\newblock In {\em Advances in neural information processing systems}, pages
  5753--5763, 2019.

\bibitem{sun2019ernie}
Yu~Sun, Shuohuan Wang, Yukun Li, Shikun Feng, Xuyi Chen, Han Zhang, Xin Tian,
  Danxiang Zhu, Hao Tian, and Hua Wu.
\newblock Ernie: Enhanced representation through knowledge integration.
\newblock {\em arXiv preprint arXiv:1904.09223}, 2019.

\bibitem{liu2019mt-dnn}
Xiaodong Liu, Pengcheng He, Weizhu Chen, and Jianfeng Gao.
\newblock Multi-task deep neural networks for natural language understanding.
\newblock In {\em Proceedings of the 57th Annual Meeting of the Association for
  Computational Linguistics}, pages 4487--4496, Florence, Italy, July 2019.
  Association for Computational Linguistics.

\bibitem{wang2018glue}
Alex Wang, Amanpreet Singh, Julian Michael, Felix Hill, Omer Levy, and Samuel
  Bowman.
\newblock {GLUE}: A multi-task benchmark and analysis platform for natural
  language understanding.
\newblock In {\em Proceedings of the 2018 {EMNLP} Workshop {B}lackbox{NLP}:
  Analyzing and Interpreting Neural Networks for {NLP}}, pages 353--355,
  Brussels, Belgium, November 2018. Association for Computational Linguistics.

\bibitem{howard2018}
Jeremy Howard and Sebastian Ruder.
\newblock Universal language model fine-tuning for text classification.
\newblock In {\em Proceedings of the 56th Annual Meeting of the Association for
  Computational Linguistics (Volume 1: Long Papers)}, pages 328--339,
  Melbourne, Australia, July 2018. Association for Computational Linguistics.

\bibitem{Sun2019}
Chi Sun, Xipeng Qiu, Yige Xu, and Xuanjing Huang.
\newblock How to fine-tune bert for text classification?
\newblock In {\em China National Conference on Chinese Computational
  Linguistics}, pages 194--206. Springer, 2019.

\bibitem{Reimers2019SBERT}
Nils Reimers and Iryna Gurevych.
\newblock Sentence-{BERT}: Sentence embeddings using {S}iamese {BERT}-networks.
\newblock In {\em Proceedings of the 2019 Conference on Empirical Methods in
  Natural Language Processing and the 9th International Joint Conference on
  Natural Language Processing (EMNLP-IJCNLP)}, pages 3982--3992, Hong Kong,
  China, November 2019. Association for Computational Linguistics.

\bibitem{SKBERT}
B.~{Wang} and C.~.~J. {Kuo}.
\newblock {SBERT-WK}: A sentence embedding method by dissecting {BERT}-based
  word models.
\newblock {\em IEEE/ACM Transactions on Audio, Speech, and Language
  Processing}, 28:2146--2157, 2020.

\bibitem{yuan2008}
T.~Yuan, D.~Moore, and A.~Grierson.
\newblock A human-computer dialogue system for educational debate: A
  computational dialectics approach.
\newblock {\em International Journal of Artificial Intelligence in Education},
  18:3--26, 2008.

\bibitem{Hunter2018APS}
Anthony Hunter.
\newblock Towards a framework for computational persuasion with applications in
  behaviour change.
\newblock {\em Argument \& Computation}, 9(1):15--40, 2018.

\bibitem{MA2020EDS}
Yukun Ma, Khanh~Linh Nguyen, Frank~Z. Xing, and Erik Cambria.
\newblock A survey on empathetic dialogue systems.
\newblock {\em Information Fusion}, 64:50--70, 2020.

\bibitem{Rakshit2019}
Geetanjali Rakshit, Kevin~K Bowden, Lena Reed, Amita Misra, and Marilyn Walker.
\newblock Debbie, the debate bot of the future.
\newblock In {\em Advanced Social Interaction with Agents}, pages 45--52.
  Springer, 2019.

\bibitem{nguyen2018dave}
Dieu~Thu Le, Cam-Tu Nguyen, and Kim~Anh Nguyen.
\newblock Dave the debater: a retrieval-based and generative argumentative
  dialogue agent.
\newblock In {\em Proceedings of the 5th Workshop on Argument Mining}, pages
  121--130, 2018.

\bibitem{Higashinaka2017}
Ryuichiro Higashinaka, Kazuki Sakai, Hiroaki Sugiyama, Hiromi Narimatsu,
  Tsunehiro Arimoto, Takaaki Fukutomi, Kiyoaki Matsui, Yusuke Ijima, Hiroaki
  Ito, Shoko Araki, et~al.
\newblock Argumentative dialogue system based on argumentation structures.
\newblock In {\em Proceedings of the 21st Workshop on the Semantics and
  Pragmatics of Dialogue}, pages 154--155, 2017.

\bibitem{rosenfeld2016persuasion}
Ariel Rosenfeld and Sarit Kraus.
\newblock Strategical argumentative agent for human persuasion.
\newblock In {\em Proceedings of the Twenty-second European Conference on
  Artificial Intelligence}, pages 320--328, 2016.

\bibitem{stab2014annotating}
Christian Stab and Iryna Gurevych.
\newblock Annotating argument components and relations in persuasive essays.
\newblock In {\em Proceedings of {COLING} 2014, the 25th International
  Conference on Computational Linguistics: Technical Papers}, pages 1501--1510,
  Dublin, Ireland, August 2014. Dublin City University and Association for
  Computational Linguistics.

\bibitem{WBAG2018}
Leila Amgoud and Jonathan Ben-Naim.
\newblock Weighted bipolar argumentation graphs: Axioms and semantics.
\newblock In {\em Proceedings of the 27th International Joint Conference on
  Artificial Intelligence}, IJCAI’18, page 5194–5198. AAAI Press, 2018.

\bibitem{Vaswani2017trans}
Ashish Vaswani, Noam Shazeer, Niki Parmar, Jakob Uszkoreit, Llion Jones,
  Aidan~N Gomez, \L~ukasz Kaiser, and Illia Polosukhin.
\newblock Attention is all you need.
\newblock In I.~Guyon, U.~V. Luxburg, S.~Bengio, H.~Wallach, R.~Fergus,
  S.~Vishwanathan, and R.~Garnett, editors, {\em Advances in Neural Information
  Processing Systems 30}, pages 5998--6008. Curran Associates, Inc., 2017.

\bibitem{wu2016google}
Yonghui Wu, Mike Schuster, Zhifeng Chen, Quoc~V Le, Mohammad Norouzi, Wolfgang
  Macherey, Maxim Krikun, Yuan Cao, Qin Gao, Klaus Macherey, et~al.
\newblock Google's neural machine translation system: Bridging the gap between
  human and machine translation.
\newblock {\em arXiv preprint arXiv:1609.08144}, 2016.

\bibitem{hochreiter1997long}
Sepp Hochreiter and J{\"u}rgen Schmidhuber.
\newblock Long short-term memory.
\newblock {\em Neural computation}, 9(8):1735--1780, 1997.

\bibitem{lin2017structured}
Zhouhan Lin, Minwei Feng, C{\'{\i}}cero~Nogueira dos Santos, Mo~Yu, Bing Xiang,
  Bowen Zhou, and Yoshua Bengio.
\newblock A structured self-attentive sentence embedding.
\newblock In {\em 5th International Conference on Learning Representations,
  {ICLR}}, 2017.

\bibitem{mikolov2013efficient}
Tomas Mikolov, Kai Chen, Greg Corrado, and Jeffrey Dean.
\newblock Efficient estimation of word representations in vector space.
\newblock {\em arXiv preprint arXiv:1301.3781}, 2013.

\bibitem{pennington2014glove}
Jeffrey Pennington, Richard Socher, and Christopher Manning.
\newblock {G}lo{V}e: Global vectors for word representation.
\newblock In {\em Proceedings of the 2014 Conference on Empirical Methods in
  Natural Language Processing ({EMNLP})}, pages 1532--1543, Doha, Qatar,
  October 2014. Association for Computational Linguistics.

\bibitem{speer2012representing}
Robyn Speer and Catherine Havasi.
\newblock Representing general relational knowledge in {C}oncept{N}et 5.
\newblock In {\em Proceedings of the Eighth International Conference on
  Language Resources and Evaluation ({LREC}'12)}, pages 3679--3686, Istanbul,
  Turkey, May 2012. European Language Resources Association (ELRA).

\bibitem{niklas2018arg}
Niklas Rach, Saskia Langhammer, Wolfgang Minker, and Stefan Ultes.
\newblock Utilizing argument mining techniques for argumentative dialogue
  systems.
\newblock In {\em 9th International Workshop on Spoken Dialogue System
  Technology}, pages 131--142. Springer, 2019.

\bibitem{aharoni2014claim}
Ehud Aharoni, Anatoly Polnarov, Tamar Lavee, Daniel Hershcovich, Ran Levy, Ruty
  Rinott, Dan Gutfreund, and Noam Slonim.
\newblock A benchmark dataset for automatic detection of claims and evidence in
  the context of controversial topics.
\newblock In {\em Proceedings of the First Workshop on Argumentation Mining},
  pages 64--68, Baltimore, Maryland, June 2014. Association for Computational
  Linguistics.

\bibitem{CoopeFarghly2020}
Samuel Coope, Tyler Farghly, Daniela Gerz, Ivan Vuli{\'c}, and Matthew
  Henderson.
\newblock {S}pan-{ConveRT}: {F}ew-shot span extraction for dialog with
  pretrained conversational representations.
\newblock In {\em Proceedings of the 58th Annual Meeting of the Association for
  Computational Linguistics}, pages 107--121, Online, July 2020. Association
  for Computational Linguistics.

\bibitem{Wolf2019HuggingFacesTS}
Thomas Wolf, Lysandre Debut, Victor Sanh, Julien Chaumond, Clement Delangue,
  Anthony Moi, Pierric Cistac, Tim Rault, Remi Louf, Morgan Funtowicz, Joe
  Davison, Sam Shleifer, Patrick von Platen, Clara Ma, Yacine Jernite, Julien
  Plu, Canwen Xu, Teven Le~Scao, Sylvain Gugger, Mariama Drame, Quentin Lhoest,
  and Alexander Rush.
\newblock Transformers: State-of-the-art natural language processing.
\newblock In {\em Proceedings of the 2020 Conference on Empirical Methods in
  Natural Language Processing: System Demonstrations}, pages 38--45, Online,
  October 2020. Association for Computational Linguistics.

\bibitem{reimers-2016-task}
Nils Reimers, Philip Beyer, and Iryna Gurevych.
\newblock Task-oriented intrinsic evaluation of semantic textual similarity.
\newblock In {\em Proceedings of {COLING} 2016, the 26th International
  Conference on Computational Linguistics: Technical Papers}, pages 87--96,
  Osaka, Japan, December 2016. The COLING 2016 Organizing Committee.

\bibitem{Wang2019}
Ran Wang, Haibo Su, Chunye Wang, Kailin Ji, and Jupeng Ding.
\newblock To tune or not to tune? how about the best of both worlds?
\newblock {\em arXiv preprint arXiv:1907.05338}, 2019.

\bibitem{wu2018Star}
Ledell~Yu Wu, Adam Fisch, Sumit Chopra, Keith Adams, Antoine Bordes, and Jason
  Weston.
\newblock Starspace: Embed all the things!
\newblock In {\em Thirty-Second AAAI Conference on Artificial Intelligence},
  2018.

\bibitem{disBERT}
Victor Sanh, Lysandre Debut, Julien Chaumond, and Thomas Wolf.
\newblock Distilbert, a distilled version of bert: smaller, faster, cheaper and
  lighter.
\newblock {\em arXiv preprint arXiv:1910.01108}, 2019.

\bibitem{roberta2018}
Yinhan Liu, Myle Ott, Naman Goyal, Jingfei Du, Mandar Joshi, Danqi Chen, Omer
  Levy, Mike Lewis, Luke Zettlemoyer, and Veselin Stoyanov.
\newblock Roberta: A robustly optimized bert pretraining approach.
\newblock {\em arXiv preprint arXiv:1907.11692}, 2019.

\bibitem{chi-improving}
Zewen Chi, Li~Dong, Bo~Zheng, Shaohan Huang, Xian-Ling Mao, Heyan Huang, and
  Furu Wei.
\newblock Improving pretrained cross-lingual language models via self-labeled
  word alignment.
\newblock In {\em Proceedings of the 59th Annual Meeting of the Association for
  Computational Linguistics and the 11th International Joint Conference on
  Natural Language Processing (Volume 1: Long Papers)}, pages 3418--3430,
  Online, 2021. Association for Computational Linguistics.

\bibitem{bergstra2012random}
James Bergstra and Yoshua Bengio.
\newblock Random search for hyper-parameter optimization.
\newblock {\em Journal of machine learning research}, 13(Feb):281--305, 2012.

\end{thebibliography}






\end{document}